\documentclass[twoside]{article}

\usepackage[accepted]{aistats2022}
\usepackage[round]{natbib}

\setcitestyle{authoryear,open={(},close={)}}

\usepackage{amssymb}
\usepackage{glossaries}
\usepackage{xcolor}
\usepackage{algorithm}
\usepackage{algpseudocode}
\usepackage{bm}

\usepackage{graphicx}
\usepackage{caption}
\usepackage{subcaption}

\usepackage{booktabs}
\usepackage{hyperref}

\newacronym{rl}{RL}{Reinforcement Learning}
\newacronym{il}{IL}{Imitation Learning}
\newacronym{drl}{DRL}{Deep Reinforcement Learning}
\newacronym{bbo}{BBO}{Black-Box Optimization}
\newacronym{ps}{PS}{Policy Search}
\newacronym{hrl}{HRL}{Hierarchical Reinforcement Learning}
\newacronym{avi}{AVI}{Approximate Value-Iteration}
\newacronym{api}{API}{Approximate Policy-Iteration}
\newacronym[plural=MDPs, firstplural=Markov Decision Processes (MDPs)]{mdp}{MDP}{Markov Decision Process}
\newacronym{dof}{DoF}{Degrees of Freedom}
\newacronym{pca}{PCA}{Principal Component Analysis}
\newacronym{rbfs}{RBFs}{Radial Basis functions}
\newacronym{promp}{ProMP}{Probabilistic Movement Primitives}
\newacronym{dmp}{DMP}{Dynamic Movement Primitives}
\newacronym{cmdp}{CMDP}{Constrained Markov Decision Processes}
\newacronym{kl}{KL}{Kullback-Leibler Divergence}
\newacronym{gae}{GAE}{Generalized Advantage Estimation}
\newacronym{trpo}{TRPO}{Trust Region Policy Optimization}
\newacronym{reps}{REPS}{Relative Entropy Policy Search}
\newacronym{creps}{CREPS}{Constrained Relative Entropy Policy Search}
\newacronym{drcreps}{DR-CREPS}{Dimensionality Reduced Constrained Relative Entropy Policy Search}
\newacronym{drreps}{DR-REPS}{Dimensionality Reduced Relative Entropy Policy Search}
\newacronym{more}{MORE}{Model-Based Relative Entropy Stochastic Search}
\newacronym{rwr}{RWR}{Reward-Weighted Regression}
\newacronym{mppca}{MPPCA}{Mixture of Probabilistic Principal Component Analysis}
\newacronym{mi}{MI}{Mutual Information}
\newacronym{pearson}{PCC}{Pearson Correlation Coefficient}
\newacronym{svd}{SVD}{Singular Value Decomposition}
\newacronym{gplvm}{GPLVM}{Gaussian Process Latent Variable Model}
\newacronym{lqr}{LQR}{Linear Quadratic Regulator}
\newacronym{pe}{PE}{Prioritized Exploration}
\newacronym{pro}{PRO}{Pearson-Correlation-Based Relevance Weighted Policy Optimization}
\newacronym{wmle}{WMLE}{Weighted Maximum Likelihood Estimate}
\newacronym{cwmle}{CWMLE}{Constrained WMLE}
\newacronym{gdr}{GDR}{Guided Dimensionality Reduction}
\newacronym{ppo}{PPO}{Proximal Policy Optimization}
\newacronym{es}{ES}{Evolution Strategies}
\newacronym{nes}{NES}{Natural Evolution Strategies}
\newacronym{nm}{NM}{Nelder-Mead Simplex}
\newacronym{lbfgs}{L-BFGS}{Limited-memory BFGS}
\newacronym{cem}{CEM}{Cross-entropy Method}

\begin{document}

% If your paper is accepted and the title of your paper is very long,
% the style will print as headings an error message. Use the following
% command to supply a shorter title of your paper so that it can be
% used as headings.
%
%\runningtitle{I use this title instead because the last one was very long}

% If your paper is accepted and the number of authors is large, the
% style will print as headings an error message. Use the following
% command to supply a shorter version of the authors names so that
% they can be used as headings (for example, use only the surnames)
%
%\runningauthor{Surname 1, Surname 2, Surname 3, ...., Surname n}

\twocolumn[

\aistatstitle{Dimensionality Reduction and Prioritized Exploration for Policy Search}

\aistatsauthor{ Marius Memmel \And Puze Liu \And  Davide Tateo \And Jan Peters }

\aistatsaddress{ TU Darmstadt \And TU Darmstadt \And TU Darmstadt \And TU Darmstadt}
]

\begin{abstract}
    Black-box policy optimization is a class of reinforcement learning algorithms that explores and updates the policies at the parameter level. This class of algorithms is widely applied in robotics with movement primitives or non-differentiable policies.
    Furthermore, these approaches are particularly relevant where exploration at the action level could cause actuator damage or other safety issues.
    However, Black-box optimization does not scale well with the increasing dimensionality of the policy, leading to high demand for samples, which are expensive to obtain in real-world systems.
    In many practical applications, policy parameters do not contribute equally to the return. Identifying the most relevant parameters allows to narrow down the exploration and speed up the learning. Furthermore, updating only the effective parameters requires fewer samples, improving the scalability of the method.
    We present a novel method to prioritize the exploration of effective parameters and cope with full covariance matrix updates. Our algorithm learns faster than recent approaches and requires fewer samples to achieve state-of-the-art results. To select the effective parameters, we consider both the Pearson correlation coefficient and the Mutual Information. 
    We showcase the capabilities of our approach on the Relative Entropy Policy Search algorithm in several simulated environments, including robotics simulations. Code is available at \href{https://git.ias.informatik.tu-darmstadt.de/ias\_code/aistats2022/dr-creps}{git.ias.informatik.tu-darmstadt.de/ias\_code/aistats2022/dr-creps}.
\end{abstract}

\section{INTRODUCTION}
\gls{bbo} is a class of Policy Search methods that tackle \gls{rl} problems in the parameter space. These methods have been widely applied to robotics applications, such as Table Tennis \citep{REPS}, Beer Pong \citep{MORE}, Pancake-Flipping \citep{kormushev2010robot}, and Juggling \citep{ConstrainedREPS}. 
By using a search distribution, it conducts the exploration at the parameter level instead of exploring at the action level. The parameter space exploration perturbs the policy parameters at the beginning of the episode instead of adding noise to actions at each time step. This approach leads to more reliable policy updates and suits better real-world learning tasks, as the generated trajectories are smoother. Furthermore,  in contrast to action level exploration, it avoids actuator damage and doesn't get low-pass filtered by the physical system~\citep{SurveyPS}. Finally, \gls{bbo} can be more suitable if the reward is not Markovian, e.g., the episodic reward only considers the maximum step reward during the trajectory.

One major challenge in \gls{bbo} is the scalability in high-dimensional tasks. The search distribution of the policy parameters is often represented as a multivariate Gaussian distribution. While the search distribution parameterized by a diagonal covariance matrix is sufficient for simple tasks, more complex tasks require correlated exploration using a full covariance matrix. Furthermore, it has been shown that a full covariance matrix produces more informative samples w.r.t. the diagonal covariance and increases the learning speed~\citep{SurveyPS}. However, the dimensionality of the full covariance matrix scales quadratically with the number of parameters. Therefore, the learning agent requires more samples to update the search distribution, because with an insufficient amount of samples the estimated covariance matrix may not be positive definite. Indeed, particularly when samples come directly from real robots, exploration is problematic in terms of time and resource costs.
% Learning the search distribution requires a dataset at the current time step consisting of samples from the distribution itself, as well as the total return achieved by executing the parameterized lower level policy in the environment, i.e., an episode. Policy search algorithms like, e.g., \gls{rwr}~\citep{RWR}, \gls{reps}~\citep{REPS} or \gls{more}~\citep{MORE}, then use the dataset to update the current policy.
% The higher dimensional the search distribution becomes, the more samples are required to update its covariance. 
% Additionally,  This growth limits the ability to learn high dimensional using a full covariance matrix as the amount of samples required for learning becomes infeasible.

To reach the optimal policy with fewer samples, we usually tackle the problem in two aspects: more informative samples or more effective policy updates. 
Noticeably, the policy parameters in a system generally do not have equal contribution to the return. 
For instance, many tasks using a 7 \gls{dof} manipulator can be solved without utilizing all \gls{dof}s. We refer to the parameters that contribute most to the return as \textsl{effective parameters} and the remaining ones as \textsl{ineffective parameters}. 
Samples spreading over the \textsl{ineffective parameters} do not provide valuable information to the learning agent, and updating the distribution w.r.t such parameters does not have an impact on the learning performance. On the contrary, if the sampling and the policy update focus on the \textsl{effective parameters}, we could obtain better learning performance.

In this paper, we propose a new \gls{bbo} algorithm, \gls{drcreps}. Our method consists of 3 components: i. \textsl{Parameter Effectiveness Metric}. To estimate the effectiveness of the parameter, we estimate the correlation measurement (e.g., \gls{pearson}, \gls{mi}) between parameters and the episodic return. ii. \textsl{Prioritized Exploration}. During the exploration process, we concentrate the sampling on the \textsl{effective} parameters by decreasing the covariance entries w.r.t the \textsl{ineffective} parameters. iii. \textsl{Guided Dimensionality Reduction}. We propose a dimensionality reduction technique for policy search algorithms that scales to the full covariance case by partially updating the search distribution w.r.t. the effective parameters. For simplicity, we first present i. for the diagonal covariance matrix case. Then, we project the full covariance matrix into a rotated space where the covariance matrix is diagonal. Subsequently, we explain ii. and iii. in the rotated space. We update the policy in the rotated space following the \gls{creps} algorithm~\citep{abdolmaleki2017deriving,ConstrainedREPS}. 
Finally, we evaluate the proposed methodology in four continuous control and simulated robotics tasks. The empirical results show the benefit of considering effective parameters for \gls{bbo}.
\vspace{-.8em}

\subsection*{Assumptions and Limitations}
In this work, we focus on learning the simple parametric policies, such as Movement Primitives and Linear Feedback Controller, under a multivariate Gaussian search distribution. While these policies can be high dimensional in the parameter space, each parameter may have a great impact on the policies' behavior. This class of policies includes many movement primitives used in robotics, e.g., \gls{dmp}~\citep{DMP} or \gls{promp}~\citep{ProMP}, linear policies, and ad-hoc, non-differentiable policies, but excludes the neural network scenario, where each parameter has no major impact on the policy.
Thus, extending the dimensionality reduction techniques presented in this work to the neural network scenario is nontrivial and requires further investigation. One possible solution in the literature is to consider specific neural network structures~\citep{choromanski2018structured}. 

In general, using a full covariance matrix will produce more informative samples but requires more samples for the policy updates. Our dimensionality reduction method allows us to generate informative samples and update policies with fewer samples simultaneously. For action space policies, e.g., neural networks, the correlation between the action dimensions is often less important than the temporal correlation, i.e., the correlation between different steps, and using the diagonal covariance of each time step does not significantly affect the performance.

Parameter exploration and \gls{bbo} approaches might be well suited for a wide variety of tasks, e.g., when the reward function is not much informative. However, classical exploration could potentially be more sample efficient in environments where the local policy changes affect the total return, e.g., standard Mujoco environments.
For this reason, we focus on specific robotics tasks where the task reward is sparse and the black box learning could be beneficial.

\subsection*{Related Work}
Dimensionality reduction is an important approach to alleviate the problem of \textit{``Curse of Dimensionality''}. It has been studied from various perspectives in \gls{il} and \gls{rl}, e.g., state space \citep{Bitzer2010, JaderbergMCSLSK17}, action space \citep{Luck2014} and parameter space \citep{Bitzer2009, Colome2014a}. In this paper, we will focus on the dimensionality reduction problem in the parameter space. 

\cite{BenAmor2012} tackle this problem by directly projecting the movement primitives into a lower-dimensional space. In their method, they use the \gls{pca} and show its application to a human grasping task. Colomé and Torras extensively studied the \gls{pca} approach in different types of movement primitives, such as \gls{dmp} \citep{DmpPCA, DmpPca2} and \gls{promp} \citep{Colome2014a, Colome2018a}. The authors showcase the method for trajectory learning with movement primitives in various applications, including bimanual manipulation of clothes. A different approach to dimensionality reduction for movement primitives is the reduction of the parameter space via a \gls{mppca} by~\cite{Tosatto2020} which makes the optimization feasible for the \gls{rl} setting. 
\cite{Bitzer2009} propose \gls{gplvm} as a non-linear dimensionality reduction technique and apply it to high dimensional \gls{dmp}s. \cite{DelgadoGuerrero2020} also utilize a \gls{gplvm} to project the data itself into a lower-dimensional space. They build a surrogate model of the reward function that they incentivize by weighing the data according to the \gls{mi} with the reward.

\cite{Ewerton2019} exploit the connection of the \gls{pearson} with the relevancy of parameters in trajectories to different objectives. They show that the computed \gls{pearson} can improve the policy optimization of trajectory distributions by reducing the exploration space. 
To go beyond the \gls{pearson} for dimensionality reduction, we also evaluate the \gls{mi} as a non-linear correlation measure. The \gls{mi} captures the information overlap between two random variables and allows us to measure how parameters influence the total return of an episode. Since its computation from samples is nontrivial, we turn to~\cite{Carrara2019} and \cite{Kraskov2004} who provide suitable estimators.
Another drawback of~\cite{Ewerton2019}'s approach is the limitation to diagonal covariance matrices that we overcome using \gls{svd} and the importance estimation of the parameters.

Some recent works focus on scaling \gls{bbo} evolutionary strategies to high dimensional parameter spaces, such as Neural Networks, by restricting the exploration space into a lower-dimensional space. \cite{Maheswaranathan2019} introduce a guided subspace in the form of a low-rank covariance matrix to reduce the search dimension, while \cite{Choromanski2019} turn to the \gls{pca} to approximate gradients in a lower-dimensional active subspace.
While both approaches are limited to linear subspaces, \cite{Sener2020} propose to use neural networks to learn the low-dimensional manifold.

\begin{figure*}
    \centering
    \includegraphics[width=1.\textwidth]{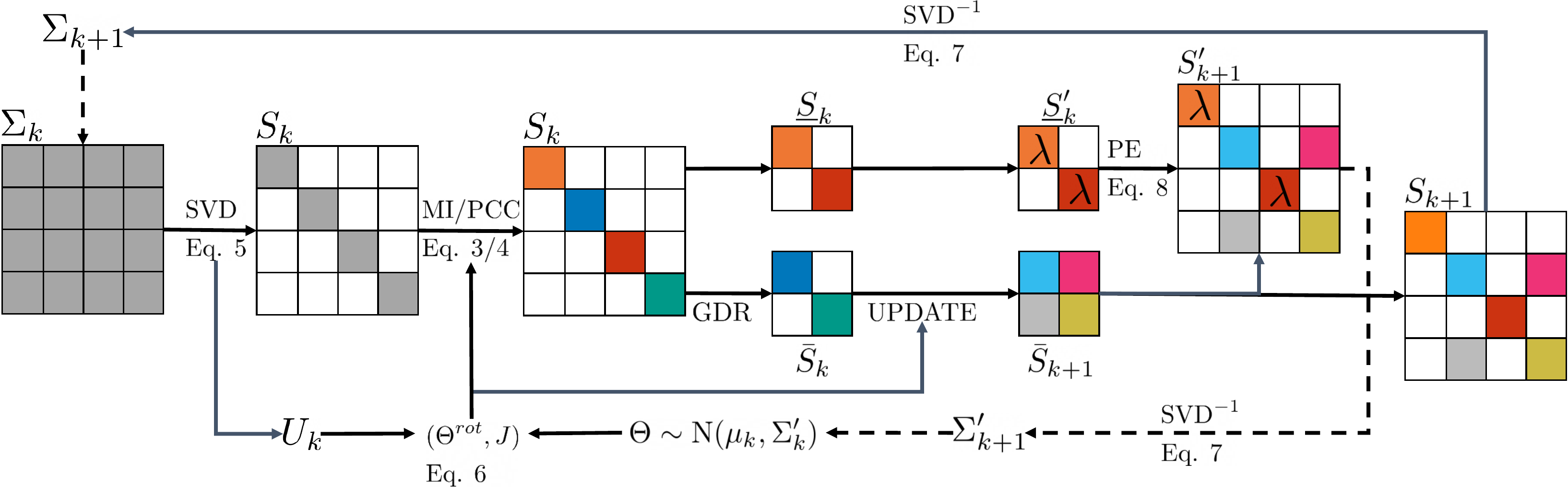}
    \caption{\acrlong{gdr} (GDR) and \acrlong{pe} (PE) on a full covariance update for a generic policy search algorithm (UPDATE).}
    \label{fig:algorithm_overview}
\end{figure*}

\section{PROPOSED METHOD}
Our method consists of three main components: First, we select the effective parameters using a correlation metric of choice. Second, we sample a new parameter vector by prioritizing exploration on the effective parameters. Third, we update the search distribution by prioritizing the important policy parameters.
We provide a step-wise visualization of our methods applied to a full covariance matrix in Fig.~\ref{fig:algorithm_overview}.

\subsection{Preliminaries}
A \gls{mdp} is defined as a $6$-tuple $\mathcal{M} = \langle\mathcal{X}, \mathcal{U}, \mathcal{P}, r, \iota, \gamma\rangle$, where $\mathcal{X}$ is the state space, $\mathcal{U}$ is the action space, $\mathcal{P}: \mathcal{X} \times \mathcal{U} \to \mathcal{X}$ is the transition  kernel, $r: \mathcal{X} \times \mathcal{U} \to \mathbb{R}$ is the reward function, $\iota: \mathcal{X} \to \mathbb{R}$ is the initial state distribution, and $\gamma \in (0,1]$ is the discount factor. 

A parametric policy $\pi_\theta$, with $\bm{\theta}\in\mathbb{R}^n$ provides the action $\bm{u}\in\mathcal{U}$ to perform in each state $\bm{x}\in\mathcal{X}$. The performance of a policy, given a parameter vector $\bm{\theta}$, is evaluated through the return, i.e., the discounted cumulative reward:
{\small
\begin{equation*}
    J(\pi_\theta) = \underset{(\bm{x}_t,\bm{u}_t) \sim \pi_\theta, \mathcal{P},\iota}{\mathbb{E}}\left[\sum_{t=0}^T \gamma^t r(\bm{x}_t,\bm{u}_t)\right],
\end{equation*}%
}%
where $T$ is the length of the trajectory.
Different to standard step-based \gls{rl}, in the \gls{bbo} setting we try to maximize the expected return $\mathcal{J}$ under a search distribution $p$:
{\small
\begin{equation*}
    \mathcal{J}(p) = \mathbb{E}_{\bm{\theta}\sim p}\left[J(\pi_\theta)\right].
\end{equation*} %
}%
In this paper, we focus on Gaussian search distributions. We define the distribution at epoch $k$ as
\begin{align*}
    p_{k}(\bm{\theta})=\mathcal{N}\left(\cdot|\bm{\mu}_{k}, \bm{\Sigma}_{k}\right),
\end{align*}
with mean $\bm{\mu}_{k}\in\mathbb{R}^{n}$, and covariance $\bm{\Sigma}_{k}\in\mathbb{R}^{n\times n}$. To be a proper probability distribution, $\bm{\Sigma}_{k}$ must be positive definite. 

A typical \gls{bbo} approach explores the environment by sampling $\bm{\theta}$ from the search distribution $p$ and evaluating the performance of the policy parameters by interacting with the environment.
After the evaluation of $N$ parameters, the algorithm updates the search distribution using the parameters matrix $\bm{\Theta}\in\mathbb{R}^{N\times n}$, which contains all the sampled parameter vectors, and the vector of returns $J(\bm{\Theta})\in\mathbb{R}^N$, containing the corresponding performance of each parameter vector.
%We define the dataset as $\mathcal{D} = \{\bm{\Theta}, \bm{J}\}=\{(\bm{\theta}_{i}, J_{i})\}_{i=1,...N}$, containing $N$ parameter samples $\bm{\theta}_{i}$ and their corresponding total return of the environment rollout $J_{i}$.
In the rest of the paper, we denote the $j$-th component of $i$-th parameter sample $\bm{\theta}_{i}$ as $\theta^{j}_{i}$.

To improve the robustness of the learning behavior and avoid premature convergence, many \gls{rl} algorithms, such as \cite{REPS} and \cite{schulman2015trust}, apply the \gls{kl} constraint to the policy update. The resulting \gls{bbo} optimization problem is
\begin{equation}
    \max_{p_k} \; \mathcal{J}(p_k), \quad \mathrm{s.t.} \;\; \mathrm{KL}\left(p_k \Vert p_{k-1} \right) \leq \epsilon.  \label{eq:problem_formulation}
\end{equation}
Using the method of Lagrangian multipliers, \gls{reps} calculates the solution for this problem as $p_{k}(\bm{\theta}) \propto p_{k-1}(\bm{\theta}) \exp( J(\pi_\theta)/\eta)$, with the temperature parameter $\eta$ automatically chosen to fulfill the \gls{kl} constraint. As the proposed update is intractable, the authors approximate the new policy $p_k$ using a Gaussian $\bm{\mu}_k, \bm{\Sigma}_k$, through sample-based \gls{wmle}. However, this results in a distribution that is no longer guaranteed to fulfill the original \gls{kl} constraint. \gls{creps}, introduced by~\cite{ConstrainedREPS}, addresses this problem by adding the \gls{kl} constraint to the \gls{wmle} fit.

The \gls{cwmle} update is formulated as:
\begin{align}
    \max_{p_{k+1}}&\, \sum^{N}_{i=1}d_{i}\log p_{k+1}\left(\bm{\theta}_{i}\right) \nonumber\\
     \text{s.t.}\, \, \, &\, \mathrm{KL}\left(p_k||p_{k+1}\right)\leq\epsilon, & \mathrm{H}\left(p_k\right)-\kappa\leq \mathrm{H}\left(p_{k+1}\right),
     \label{eq:constrained_reps}
\end{align}
with the search distribution $p_{k}$ at epoch $k$, the weight $d_i$ computed by the \gls{reps} algorithm corresponding to the parameter sample $\bm{\theta}_i$, the \gls{kl} bound $\epsilon$, and the maximally allowed entropy decrease $\kappa$.

\subsection{Estimating Effective Parameters}
To determine the \textsl{effective parameters} and conversely the \textsl{ineffective parameters}, we need a parameter effectiveness metric to measure the influence of the policy parameters on the episodic return. Several metrics are proposed in the pioneering work, such as \acrlong{pearson} for policy updates~\citep{Ewerton2019} and \acrlong{mi} (referred to as PIC~\citep{Furuta2021}) for the task complexity evaluation. 

\textbf{\glsdesc{pearson}} measures the linear relationship between two random variables that can be estimated directly with samples. The coefficient is bound by $[+1, -1]$ for a fully linear positive and negative correlation respectively. We measure the \gls{pearson} of the $j$-th component $\bm{\Theta}^j$ with the cumulative return $\bm{J}$ based on the samples and use the absolute value as the correlation measure
{\small
\begin{equation}\label{eq:pearsonr}
    C_{PCC}[\bm{\Theta}^j;\bm{J}] = \frac{ \left|\sum_i\left(\theta^j_i-\hat{\theta} \right )\left(J_i-\hat{J} \right ) \right|}{\sqrt{\sum_i\left(\theta_i^j-\hat{\theta}^j \right )^2 \sum_i\left(J_i-\hat{J} \right )^2}},
\end{equation} %
}%
where $\hat{\theta}^{j}$ and $\hat{J}$ are the respective means. 

\textbf{\glsdesc{mi}} possesses an attractive information-theoretic interpretation and can be used to measure non-linear correlations. In theory, \gls{mi} can capture a more complex relationship between two random variables than \gls{pearson}. On the flipside, estimating it directly from samples is difficult, since it is defined in terms of probability densities over the random variables which require a density estimation from samples. The \gls{mi} between the $j$-th policy parameter and the return is
{\small
\begin{equation}\label{eq:mi}
    C_{MI}[\bm{\Theta}^j;\bm{J}] = \int \; p\left(\bm{\theta}^j,\bm{J} \right )\; \log \frac{p\left(\bm{\theta}^j, \bm{J} \right )}{p\left(\bm{\theta}^j \right )p\left(\bm{J} \right )}d\bm{\theta}^j d\bm{J}.
\end{equation} %
}%
Since a sample-based estimation of the \gls{mi} is harder and has higher variance than the \gls{pearson}, we provide a detailed comparison of different estimators in Sec.~\ref{sec:mutual_info_estimation} of the Supplementary Materials.

After measuring the correlation between the parameters and the return, we select a fixed number $m$ of parameters with the highest correlation and assign them to the effective class $\bar{\bm{\varepsilon}}$. The remaining parameters are considered as the ineffective class $\underline{\bm{\varepsilon}}$.

\subsection{The Full Covariance Case}
\label{subsec:full_cov}
As presented by \cite{Ewerton2019}, finding the corresponding effectiveness or relevance of the parameters is trivial for the diagonal covariance case. However, this approach neglects the inter-dimensional dependencies and leads to limitations in the policy complexity and the learning efficiency. We, therefore, consider the full covariance case and propose an algorithm to overcome these limitations.

Identifying the effectiveness of the parameters in the full covariance matrix becomes nontrivial due to the relationships between them. To achieve independence among the parameters, we transform the covariance matrix into a space where every dimension is uncorrelated. This can be achieved using the \gls{svd}, which decomposes the covariance matrix into two rotation matrices $\bm{U}$ and $\bm{V}$ as well as a diagonal matrix $\bm{S}$ with the singular values on the diagonal. 
More formally, decomposing covariance matrix $\bm{\Sigma}_{k}$ gives us
\begin{equation}
    \bm{\Sigma}_{k}=\bm{U}_k\bm{S}_{k}\bm{V}_{k}^T,
\end{equation}
with $\bm{U}_{k},\bm{S}_{k},\bm{V}_{k}\in\mathbb{R}^{n\times n}$.
$\bm{\Sigma}_k$ is a positive definite symmetric matrix, therefore we have $\bm{U}_k=\bm{V}_k$.

We then define a policy in the rotated space as
\begin{equation*}
    p^{rot}_{k}=\mathcal{N}\left(\cdot|\bm{\mu}^\text{rot}_k, \bm{S}_{k}\right),
\end{equation*}
with $\bm{\mu}^\text{rot}_k=\bm{0}\in\mathbb{R}^n$, $\bm{S}_{k}=\text{diag}\left(\sigma^2_{0}, ..., \sigma^2_{n}\right)\in\mathbb{R}^{n\times n}$.
Since the parameters are no longer correlated in the rotated space, it again becomes trivial to identify the covariance entries corresponding to the effective parameters.

To estimate the contribution of the parameters to the total return, we must also rotate each parameter sample $\bm{\theta}$ by
\begin{equation}\label{eq:sampel_transform}
    \bm{\theta}^{rot}= \bm{U}_k^T\left(\bm{\theta} - \bm{\mu}_k\right),
\end{equation}
which projects them into the rotated space of $\bm{S}_{k}$. We use the projected samples to compute the correlation between the diagonal of $\bm{S}_{k}$ and $\bm{J}(\bm{\Theta})$.

The introduced transformations allow us to work on a rotated policy parameterized by a diagonal covariance matrix. Note that we have to recompute the \gls{svd} at every epoch since updating $\bm{S}_{k}$ (in Sec. \ref{subsec:guided_dim_red}) can lead to a non-diagonal structure. Through the re-computation, we guarantee a diagonal structure of the covariance matrix before every modification or update of it. Assuming an update of $(\bm{\mu}_{k}^{rot}, \bm{S}_{k})$ returning $(\bm{\mu}_{k+1}^{rot}, \bm{S}_{k+1})$, we must project the rotated policy $p_{k+1}^{rot}$ back to the original space for the policy evaluations.
We utilize the previously computed $\bm{U}_{k}$ to project back the updated mean and the updated covariance matrix
\begin{equation} \label{eq:dist_transform}
    \bm{\mu}_{k+1} = \bm{\mu}_k + \bm{U}_{k}\bm{\mu}^\text{rot}_{k+1}, \quad \bm{\Sigma}_{k+1}=\bm{U}_{k}\bm{S}_{k+1}\bm{U}_{k}^T,
\end{equation}
yielding our new policy $p_{k+1}$.

\subsection{Prioritized Exploration}
\label{subsec:prioritized_exploration}
In the following paragraph, we describe our \gls{pe} approach.
Having identified the contribution to the total return of each parameter in the rotated space, we split them up into sets of effective $\bar{\bm{\varepsilon}}$ and ineffective parameters $\underline{\bm{\varepsilon}}$ with fixed size each. Both sets $\bar{\bm{\varepsilon}}, \underline{\bm{\varepsilon}}$ capture the effectiveness of the parameters in the rotated space, i.e., similarity between $\bm{\Theta}^\text{rot}$ and $\bm{J}(\bm{\Theta})$ corresponding to the elements on the diagonal of $\bm{S}$. Without loss of generality, we decompose the total diagonal covariance matrix $\bm{S}$ into covariance matrices w.r.t. effective parameters and ineffective parameters denoted as $\bar{\bm{S}}\in \mathbb{R}^{m\times m}, \underline{\bm{S}}\in \mathbb{R}^{(n-m) \times (n-m)}$, respectively.
We use this separation to prioritize the exploration on the effective parameters by reducing exploration to the ineffective ones.
Therefore, we scale the ineffective covariance $\underline{\bm{S}}$ by hyperparameter $\lambda \in (0, 1)$ during the sampling process described as
\begin{align}\label{eq:mod_cov}
    \underline{\bm{S}}' = \lambda\underline{\bm{S}}.
\end{align}
Then, we substitute the covariance matrix corresponding to the ineffective parameter $\underline{\bm{S}}$ by the modified one $\underline{\bm{S}}'$, i.e. $\bm{S}' = \textproc{subst}(\bm{S}, \underline{\bm{S}}', \underline{\bm{\varepsilon}})$.
The \textproc{subst} function is simply placing the modified entries of $\underline{\bm{S}}'$ at their original position in $\bm{S}$ as visualized in Fig.~\ref{fig:algorithm_overview}.

After the substitution, sampling $\bm{\Theta}$ involves the modified search distribution $\mathcal{N}(\cdot|\bm{\mu}, \bm{S}')$ parameterized by the modified covariance matrix $\bm{S}'$ and the mean $\bm{\mu}$ of the true search distribution $\mathcal{N}(\cdot|\bm{\mu},\bm{S})$. After obtaining the dataset, we update the policy with an arbitrary policy search algorithm. Note that we estimate the parameter effectiveness before each sampling step. Therefore, we only use the modified $\bm{S}'$ for the sampling process and always update the unmodified $\bm{S}$. Otherwise, the covariance would shrink over time since $\lambda \in (0, 1)$.

\subsection{Guided Dimensionality Reduction}
\label{subsec:guided_dim_red}
To tackle the scalability issue of policy search algorithms, we further exploit our estimation of effective parameters $\bar{\bm{\varepsilon}}$ to introduce \gls{gdr}. Congruent with our intuition about the varying contribution of parameters to the total return, we now only update the effective ones that promise to increase the total return the most. We again focus on the rotated space and the transformation described in Sec.~\ref{subsec:full_cov}.

Running a full covariance policy search update on $\bar{\bm{S}}_{k}$ with $\bm{\Theta}^{rot}$, and $J(\bm{\Theta})$ gives us the updated covariance matrix $\bar{\bm{S}}_{k+1}\in\mathbb{R}^{m\times m}$. Note that $\bar{\bm{S}}_{k+1}$ is no longer guaranteed to be diagonal due to the full covariance update. Then, we compose the new covariance matrix $\bm{S}_{k+1}$ in the rotated space by substituting the entries of $\bar{\bm{S}}_{k}$ with $\bar{\bm{S}}_{k+1}$, i.e., $\bm{S}_{k+1} = \textproc{subst}(\bm{S}_{k}, \bar{\bm{S}}_{k+1}, \bar{\bm{\varepsilon}})$. 
We can then project $\bm{S}_{k+1}$ back into the original space via an inversion of the \gls{svd} resulting in $\bm{\Sigma}_{k+1}$.
Analogously, we update $\bar{\bm{\mu}}_k^\text{rot}=\bm{0}$ corresponding to the effective parameters to obtain $\bar{\bm{\mu}}_{k+1}^\text{rot}$. Substituting the updated values back into $\bm{\mu}_k^{rot}$ we get $\bm{\mu}_{k+1}^{rot} = \textproc{subst}(\bm{\mu}_k^{rot}, \bar{\bm{\mu}}^{rot}_{k+1}, \bar{\bm{\varepsilon}})$. Projecting it back to the original space we yield $\bm{\mu}_{k+1}$.

Since we choose $m\ll n$, updating the covariance matrix in the rotated space requires fewer samples than in the original space. Updating only the effective parameters lets us make the most efficient use of these samples without sacrificing performance. 
Reducing the number of samples makes it viable to use a full covariance matrix, even for high dimensional tasks, assuming an appropriate number of effective parameters $m$ is chosen.
Furthermore, this approach complements \gls{pe} perfectly since it maintains a diagonal structure of the ineffective parameters. Assuming similar estimations of the parameter effectiveness at each epoch, the samples would not contain much useful information to update the ineffective parameters as a result of the reduced exploration in exactly those.  

\subsection{Practical Implementation}
\begin{algorithm}[t]
\caption{\gls{drcreps}}\label{alg:c_reps_red}
{
\begin{algorithmic}[1]
\Procedure{update}{$\bm{\mu}_k,\bm{\Sigma}_k,\bm{\Theta},\bm{J}$}
    \State $\bm{U_k}, \bm{S_k} \gets \text{SVD}(\bm{\Sigma}_k)$
    \State $\bm{\Theta}^\text{rot}\leftarrow \Call{sampleProj}{\bm{U}_k, \bm{\mu}_k, \bm{\Theta}}$, Eq. \eqref{eq:sampel_transform}
    \State $\bm{C} \gets \Call{correlation}{\bm{\Theta}^\text{rot}, \bm{J}}$, Eq. \eqref{eq:pearsonr} or \eqref{eq:mi}
    \State $\bar{\bm{\varepsilon}}, \underline{\bm{\varepsilon}} \gets \Call{identifyEffectiveParameters}{\bm{C}}$
    \State $\eta^{*} \gets \Call{minimizeDualFunction}{\bm{\Theta}^\text{rot}, \bm{J}}$
    \State $\bm{d}_\theta \gets \exp\left(\frac{\bm{J} - \max \bm{J}}{\eta^{*}}\right)$
    \State $\bm{\mu}_{k+1},\bm{\Sigma}_{k+1} \gets$ $\Call{drCwmle}{\bm{U}_k, \bm{S}_k, d_\theta, \bar{\bm{\varepsilon}}}$
%   \State $\mu_{k+1},\Sigma_{k+1} \gets$$\Call{drCwmle}{U_k, S_k,  \theta', d_\theta, \bm{\varepsilon}}$
    \State \Return $\bm{\mu}_{k+1},\bm{\Sigma}_{k+1}$
\EndProcedure
\Statex
\Procedure{drCwmle}{$\bm{U}_k, \bm{S}_k, d_\theta, \bar{\bm{\varepsilon}}$}
    \State $\bar{\bm{\mu}}_k^{rot},\bar{\bm{S}}_k \gets \Call{getEffectiveDist}{\bm{S}_k, \bar{\bm{\varepsilon}}}$
    \State $\bar{\bm{\mu}}_{k+1}^{rot},\bar{\bm{S}}_{k+1} \gets \Call{cwmle}{\bar{\bm{\mu}}_k^{rot}\bar{\bm{S}}_k, d_\theta}$
    \State $\bm{\mu}_{k+1}^{rot} \gets \Call{subst}{\bm{\mu}_{k}^{rot}, \bar{{\bm{\mu}}}{}_{k+1}^{rot}, \bar{\bm{\varepsilon}}}$
    \State $\bm{S}_{k+1} \gets \Call{subst}{\bm{S}_{k}, \bar{\bm{S}}{}_{k+1}, \bar{\bm{\varepsilon}}}$
    \State $\bm{\mu}_{k+1}, \bm{\Sigma}_{k+1} \gets \Call{backProj}{\bm{\mu}_{k+1}^{rot}, \bm{S}_{k+1}}$, 
    \Statex \Comment{Eq. \eqref{eq:dist_transform}}
    %\State $\bm{\Theta}, \bm{J} \gets$$\Call{PE}{\bm{\mu}_{k+1}, \bm{U}_k,  \bm{S}_{k+1}, \bm{\varepsilon}}$
    \State \Return $\bm{\mu}_{k+1},\bm{\Sigma}_{k+1}$
\EndProcedure
\Statex
\Procedure{samplePE}{$\bm{U}_k$, $\bm{S}_{k+1}$, $\underline{\bm{\varepsilon}}$}
    \State $\underline{\bm{S}}{}_{k+1}' \gets \lambda \underline{\bm{S}}{}_{k+1}$, Eq. \eqref{eq:mod_cov}
    \State $\bm{S}'_{k+1} \gets \Call{subst}{\bm{S}_{k+1}, \underline{\bm{S}}{}_{k+1}', \underline{\bm{\varepsilon}}}$
    \State $\bm{\Sigma}_{k+1}' \gets \bm{U}_k \bm{S}'_{k+1} \bm{U}_k^T$
    \State $\bm{\Theta} \sim \mathcal{N}(\cdot|\bm{\mu}_{k+1}, \bm{\Sigma}_{k+1}')$ 
    \State $\bm{J} \gets \mathrm{Rollout}(\bm{\Theta})$
    \State \Return $\bm{\Theta}$, $\bm{J}$
\EndProcedure
\end{algorithmic}
}
\end{algorithm}
The proposed techniques can be applied to a wide variety of \gls{bbo} algorithms. In Alg.~\ref{alg:c_reps_red} we show how the dimensionality reduction for the full covariance case can be adapted to the \gls{creps} algorithm. In Fig.~\ref{fig:algorithm_overview} we provide a visual intuition of the algorithm update and sampling technique.
For simplicity, we omit the hyperparameters $\epsilon$ and $\kappa$, i.e. the \gls{kl} bound and entropy decrease bound parameters, needed to solve the optimization problems described in Eq.~\eqref{eq:problem_formulation} and Eq.~\eqref{eq:constrained_reps}.

Each update step starts by projecting the full covariance matrix into a space where all dimensions are uncorrelated using \gls{svd}. We also rotate the parameter samples $\bm{\Theta}$ into that space by applying the obtained square rotation matrix $\bm{U}_k$.
The procedure \textproc{correlaton} uses the rotated samples $\bm{\Theta^{rot}}$ to compute the $\bm{C}_{MI}/ \bm{C}_{PCC}$ of each parameter w.r.t. $\bm{J}$. The procedure \textproc{identifyEffectiveParameters} determines the set of effective parameters $\bar{\bm{\varepsilon}}$ and ineffective parameters $\underline{\bm{\varepsilon}}$.
To obtain the optimal value of the Lagrangian multiplier $\eta^{*}$, we minimize the dual function of \gls{reps} and use $\eta^{*}$ as temperature parameter to compute the weight vector $\bm{d}_\theta$ for the \gls{cwmle}.

To perform \gls{cwmle} on the reduced dimensionality, we first call the procedure \textproc{getEffectiveDistribution} to create a new diagonal distribution $\mathcal{N}(\cdot|\bar{\bm{\mu}}{}_k^{rot}, \bar{\bm{S_k}})$ based on the effective parameters $\bar{\bm{\varepsilon}}$. We then apply the policy search update using the previously computed weights $\bm{d}_\theta$. Next, we substitute the updated distribution parameters $(\bar{\bm{\mu}}{}_{k+1}^{rot}, \bar{\bm{S}}_{k+1})$ back to obtain the $\mathcal{N}(\cdot|\bm{\mu}_{k+1}^{rot}, \bm{S}_{k+1})$. Finally, we rotate the resulting distribution back to the original space yielding a new search distribution $p_{k+1}(\cdot)=\mathcal{N}(\cdot|\bm{\mu}_{k+1}, \bm{\Sigma}_{k+1})$.

We apply \gls{pe} by multiplying $\underline{\bm{S}}{}_{k+1}$ with $\lambda$ before the substitution into $\bm{S}{}_{k+1}$. Since $\bar{\bm{S}}$ and $\underline{\bm{S}}$ are uncorrelated in the rotated space, scaling $\underline{\bm{S}}$ by $\lambda$ will not affect $\bar{\bm{S}}$. We rotate the sampling covariance $\bm{S}'_{k+1}$ back to the original space via $\bm{\Sigma}'_{k+1} = \bm{U}_k \bm{S}_{k+1}' \bm{U}_k^{T}$ and obtain the prioritized samples from $\mathcal{N}(\cdot|\bm{\mu}_{k+1}, \bm{\Sigma}_{k+1}')$.
As described in Sec.~\ref{subsec:prioritized_exploration}, we only use $\bm{\Sigma}'_{k+1}$ for the sampling process, while we still consider $\bm{\Sigma}_{k+1}$ as target distribution for the \gls{kl} and entropy bounds.

\section{EXPERIMENTAL EVALUATION}

In this section, we evaluate the proposed method in four different environments. All our implementations are based on the MushroomRL library~\citep{mushroom_rl}. In all experiments, we use 25 random seeds for evaluation. For each learning curve, we plot the mean and 95\% confidence interval. We set a fixed number of episodes for each epoch. The performance is evaluated at the end of each epoch by sampling parameters from the current search distribution.

\begin{figure}[b]
    % \vspace{-1em}
     \centering
     \begin{subfigure}[b]{0.15\textwidth}
         \centering
         \includegraphics[width=2.2cm, height=2.5cm]{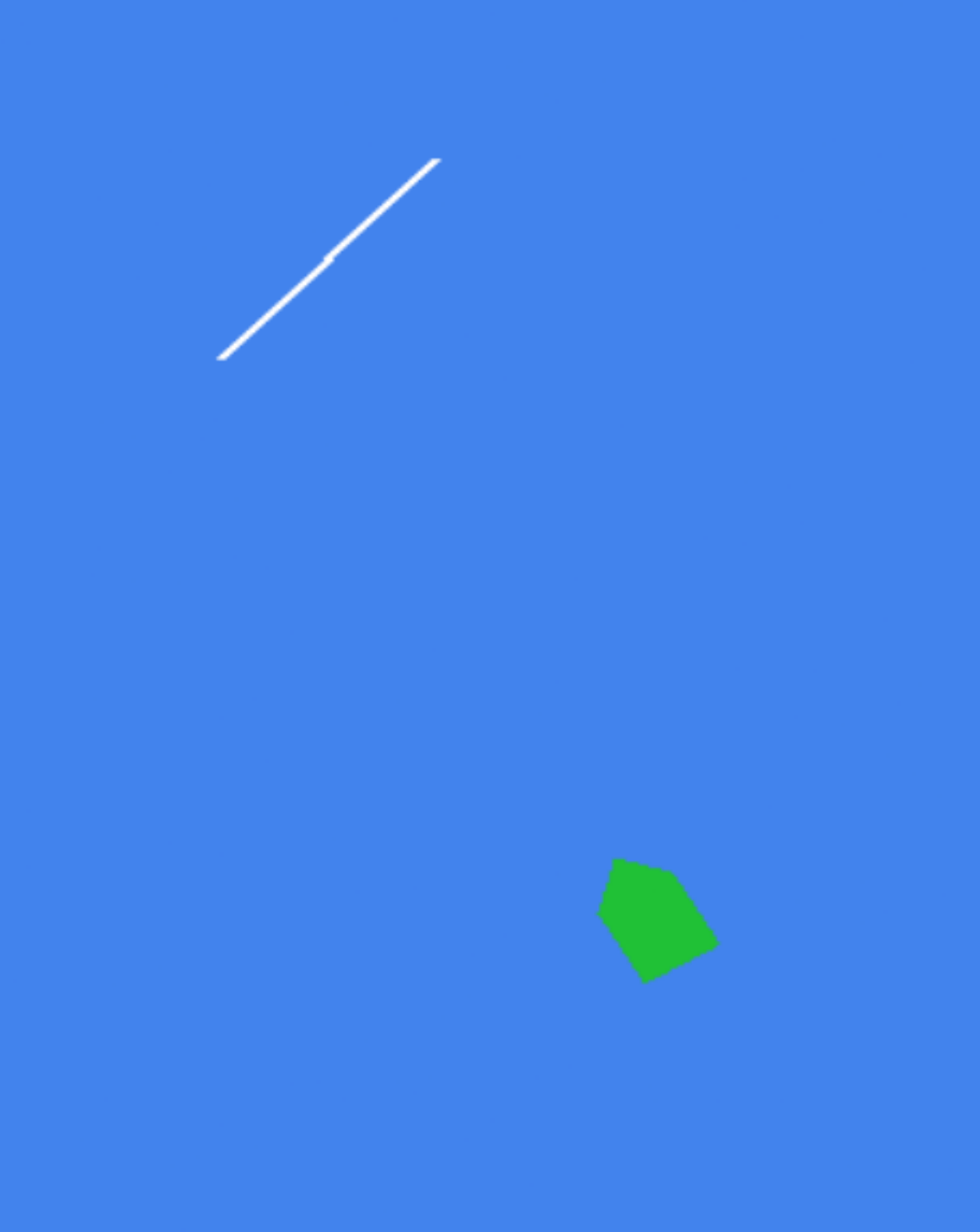}
         \caption{ShipSteering}
         \label{fig:ship_steering}
     \end{subfigure}
    %  \hfill
     \centering
     \begin{subfigure}[b]{0.15\textwidth}
         \centering
         \includegraphics[width=2.2cm, height=2.5cm]{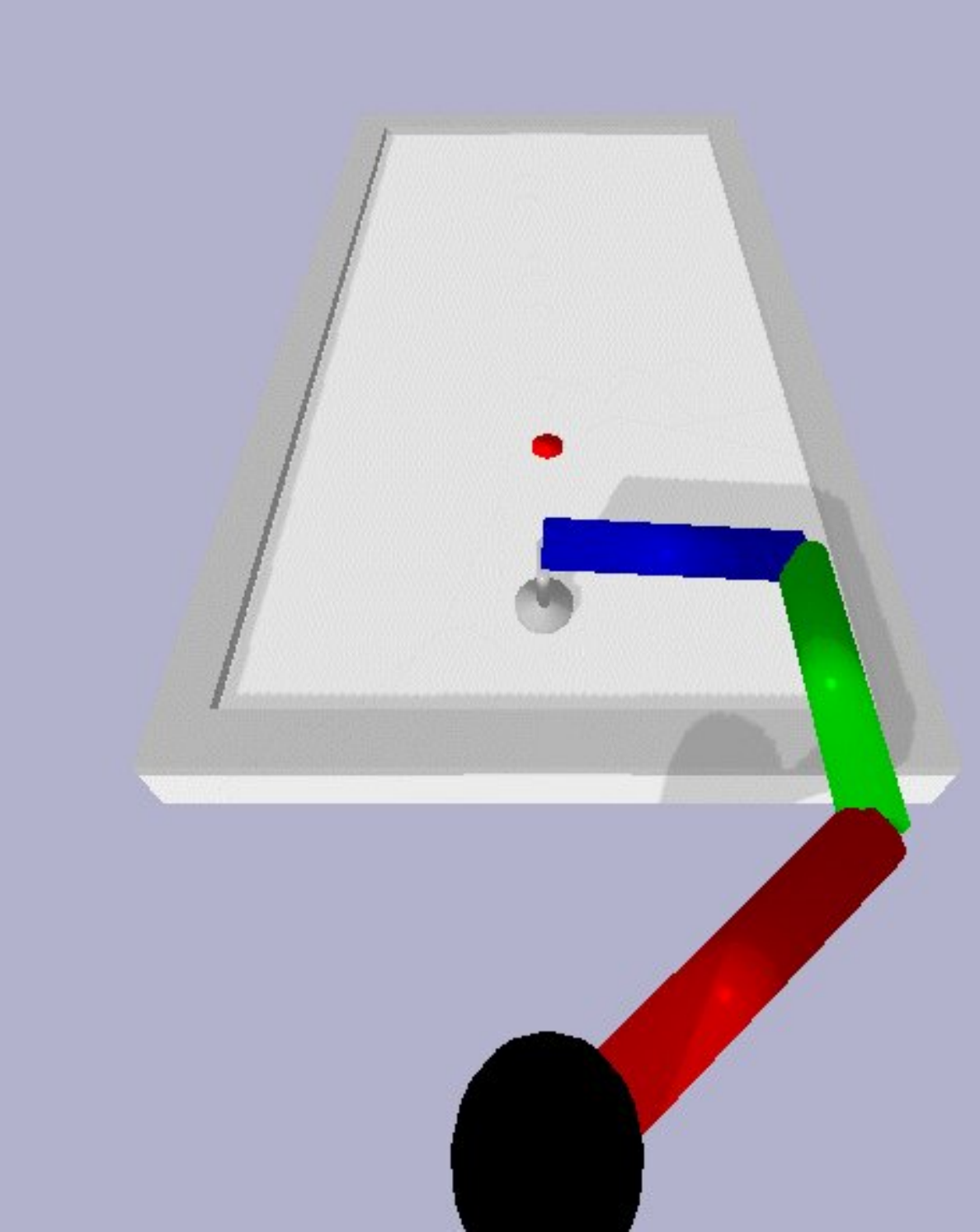}
         \caption{AirHockey}
         \label{fig:air_hockey}
     \end{subfigure}
    %  \hfill
     \begin{subfigure}[b]{0.15\textwidth}
         \centering
         \includegraphics[width=2.5cm, height=2.5cm]{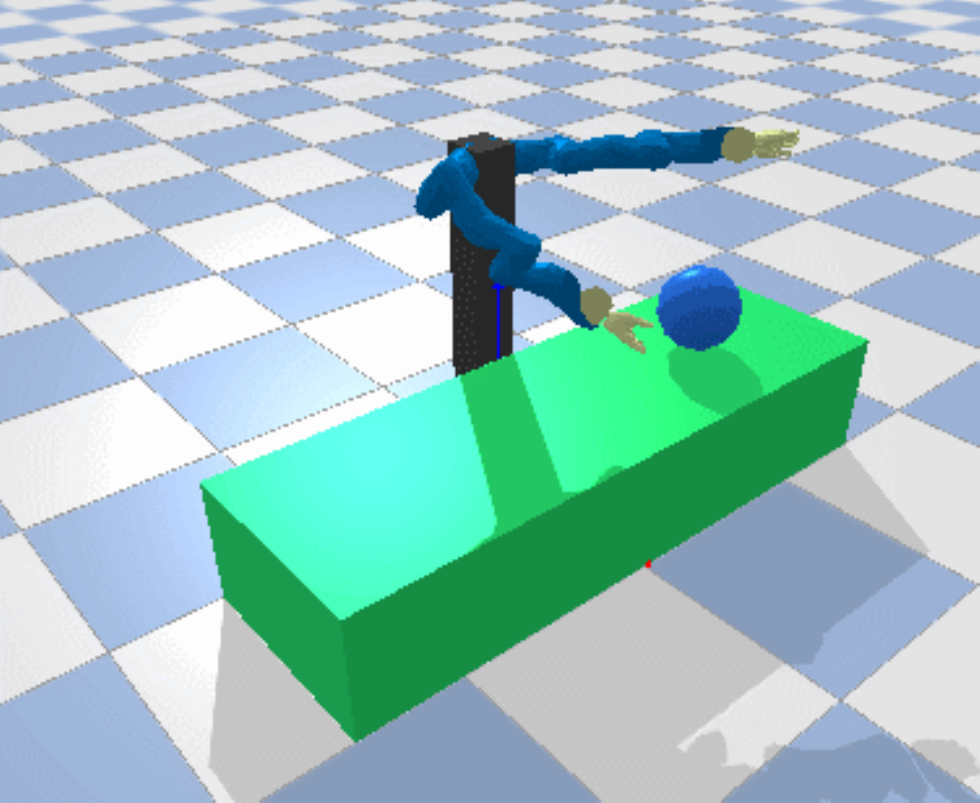}
         \caption{BallStopping}
         \label{fig:ball_stopping}
     \end{subfigure}
        \caption{Test environments}
        \label{fig:env_ship_hockey_ball}
\end{figure}

First, we test our algorithm in a modified LQR problem where only a part of the controller parameters is effective.
We construct a $10$-dimensional LQR corresponding to a $10\times10$ weight matrix of the linear regressor, i.e., $100$ parameters. The effective parameters are the non-zero entries of the optimal gain matrix which we can compute iteratively.
Then, we demonstrate several robotic simulations including \textsl{ShipSteering} (steering a ship through a gate), \textsl{AirHockey} (shooting a goal in a game of air hockey) described in \citep{CORL_2021_Air_Hockey}, and \textsl{BallStopping} (stopping a ball from rolling off a table) visualized in Fig.~\ref{fig:ship_steering}, Fig.~\ref{fig:air_hockey}, and Fig.~\ref{fig:ball_stopping}, respectively. 

We use a multivariate Gaussian distribution as search distribution and a deterministic linear policy. This policy can be a linear controller (\textit{LQR}), a CMAC approximator~\citep{albus1975new} with rectangular tiles (\textit{ShipSteering}), or a \gls{promp} (\textit{AirHockey}, \textit{BallStopping}).

We compare our approach mainly to \gls{bbo} approaches. As the main baseline, we select \gls{creps} and \gls{more}~\citep{MORE} representing policy search algorithms shown to outperform \gls{rwr} and \gls{reps}. 
We include in our comparison also two gradient-based deep actor-critic methods, \gls{trpo}~\citep{schulman2015trust} and \gls{ppo}~\citep{schulman2017proximal}, using neural policies. Lastly, we also compare our method with \gls{nes}~\citep{wierstra2014natural}, an evolutionary search strategy. Additional baselines can be found in the appendix.

The Supplementary Materials provides detailed information on the environmental setup in Sec.~\ref{sec:environment_descriptions}, hyperparameter sweep in Sec.~\ref{sec:experimental_details}, and an ablation study on the number of selected parameters $m$ and $\lambda$ in Sec.~\ref{sec:influence_lambda_m}.

\begin{figure}[t]
     \centering
     \includegraphics[width=0.47\textwidth]{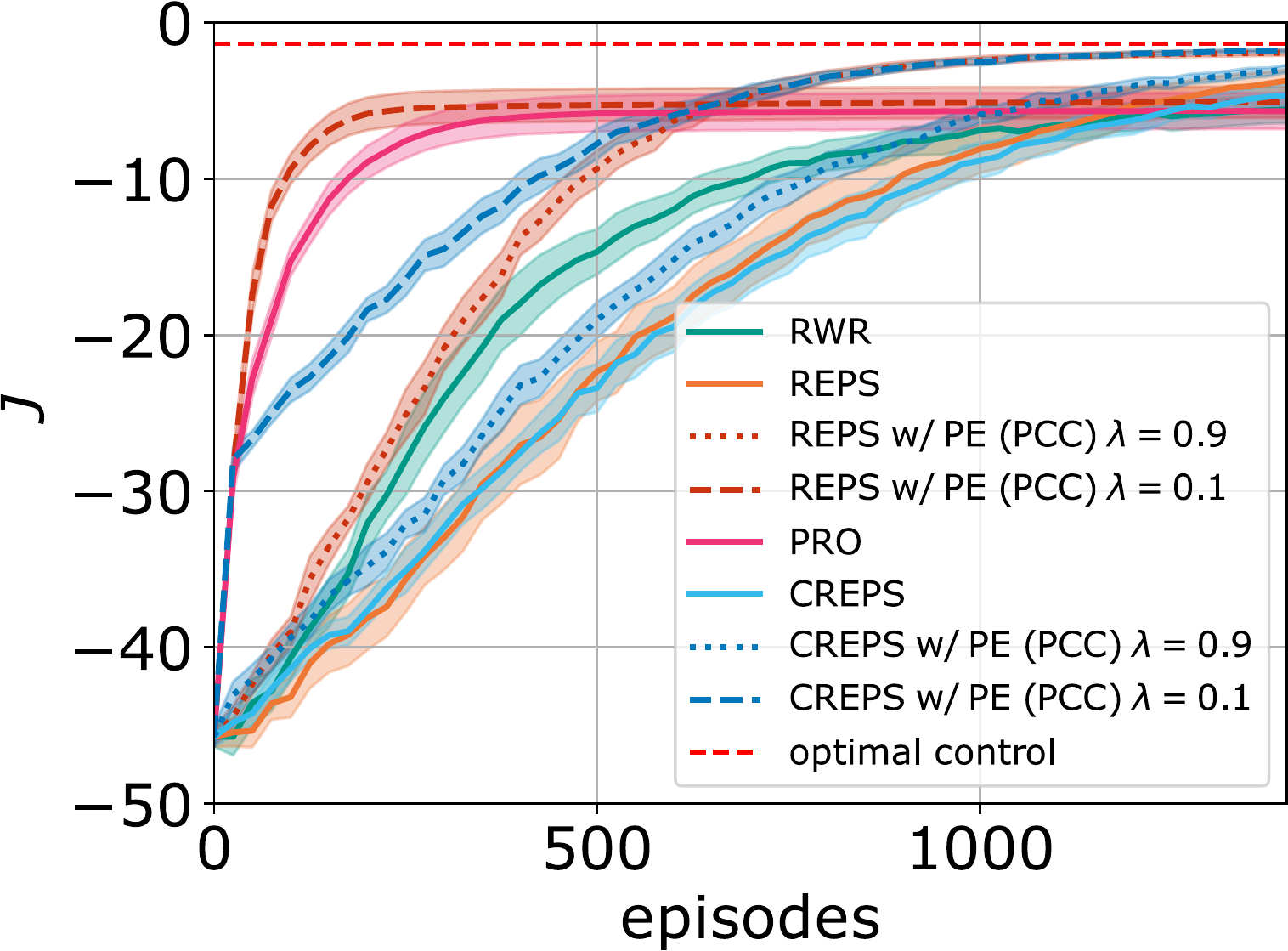}
     \caption{LQR - DiagCov}
     \label{fig:lqr_diag}
    %  \vspace{-1em}
 \end{figure}
 
\subsection{Effect of Prioritized Exploration}
To better analyze the behavior of our approach, we introduce a 10-dimensional \gls{lqr} environment characterized by three effective and seven ineffective dimensions.

We begin our evaluation on the modified \gls{lqr} and use the diagonal covariance matrix to showcase the performance of \gls{pe} without \gls{gdr}.
In Fig.~\ref{fig:lqr_diag}, \gls{pe} shows significantly faster learning when applied to both \gls{reps} and \gls{creps}. Especially \gls{reps} with \gls{pe} and $\lambda=0.1$ profits from the focus on the effective parameters. Reducing the ineffective parameters' covariance to a mere 10\% causes the learning curve to jump during the initial epochs. This is a clear sign that the effective parameters were leveraged. However, for \gls{reps} this behavior is too greedy, causing premature convergence as also experienced by \gls{pro} \citep{Ewerton2019} and \gls{rwr} \citep{SurveyPS}. Tuning hyperparameter $\lambda$ to a higher value allows for some exploration in the other parameters compensating potential mistakes in the parameter identification which would otherwise cause a too greedy behavior. Choosing $\lambda=0.9$ makes \gls{reps} with \gls{pe} converge to the optimal policy while still outperforming the vanilla version. Note that \gls{creps} with \gls{pe} does not suffer from this issue due to the additional entropy constraint designed to counteract an excessive greedy behavior, avoiding premature convergence. %We provide a more detailed ablation study of hyperparameter $\lambda$ in Sec.~\ref{sec:influence_lambda_m} of the Supplementary Materials.

\subsection{Effect of Dimensionality Reduction with Full Covariance Matrix}
 \begin{figure*}[t]
    \begin{minipage}[t]{0.32\textwidth}
        \centering
         \includegraphics[width=\textwidth]{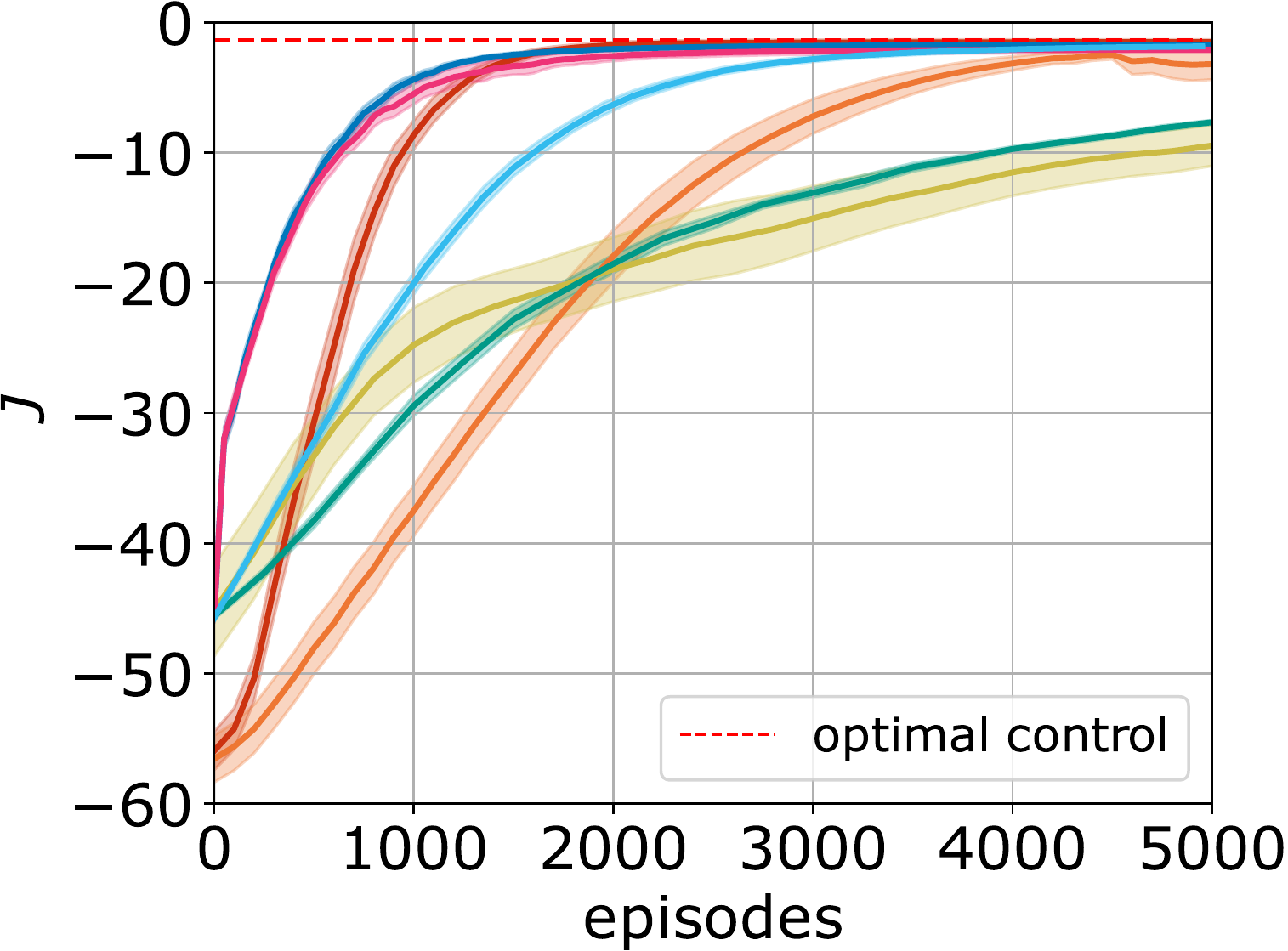}
         \caption{LQR - FullCov}
         \label{fig:lqr_full}
    \end{minipage}
    \begin{minipage}[t]{0.66\textwidth}
     \centering
     \begin{subfigure}[t]{0.49\textwidth}
         \centering
         \includegraphics[width=\textwidth]{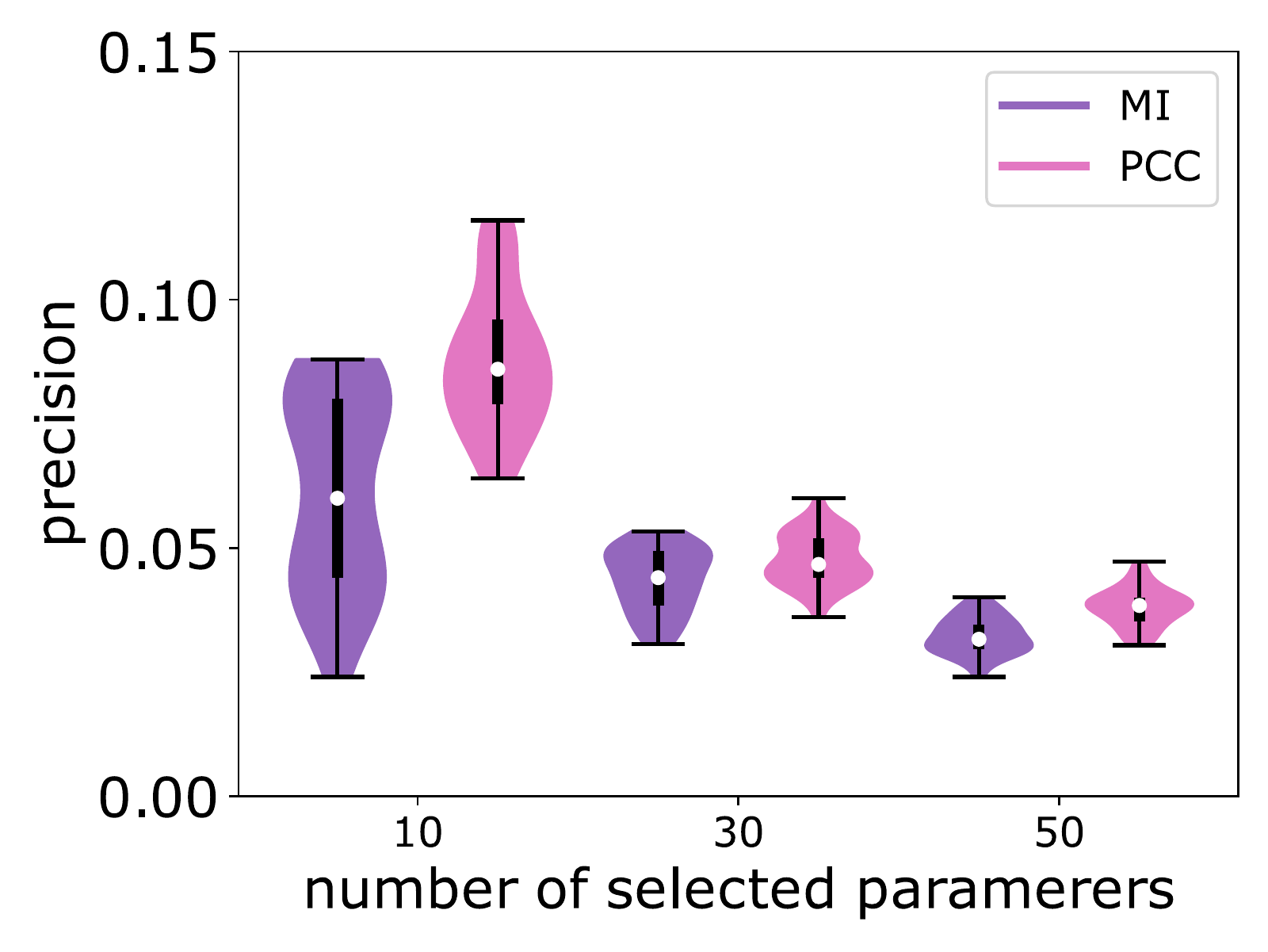}
     \end{subfigure}
     \hfill
     \begin{subfigure}[t]{0.49\textwidth}
         \centering
         \includegraphics[width=\textwidth]{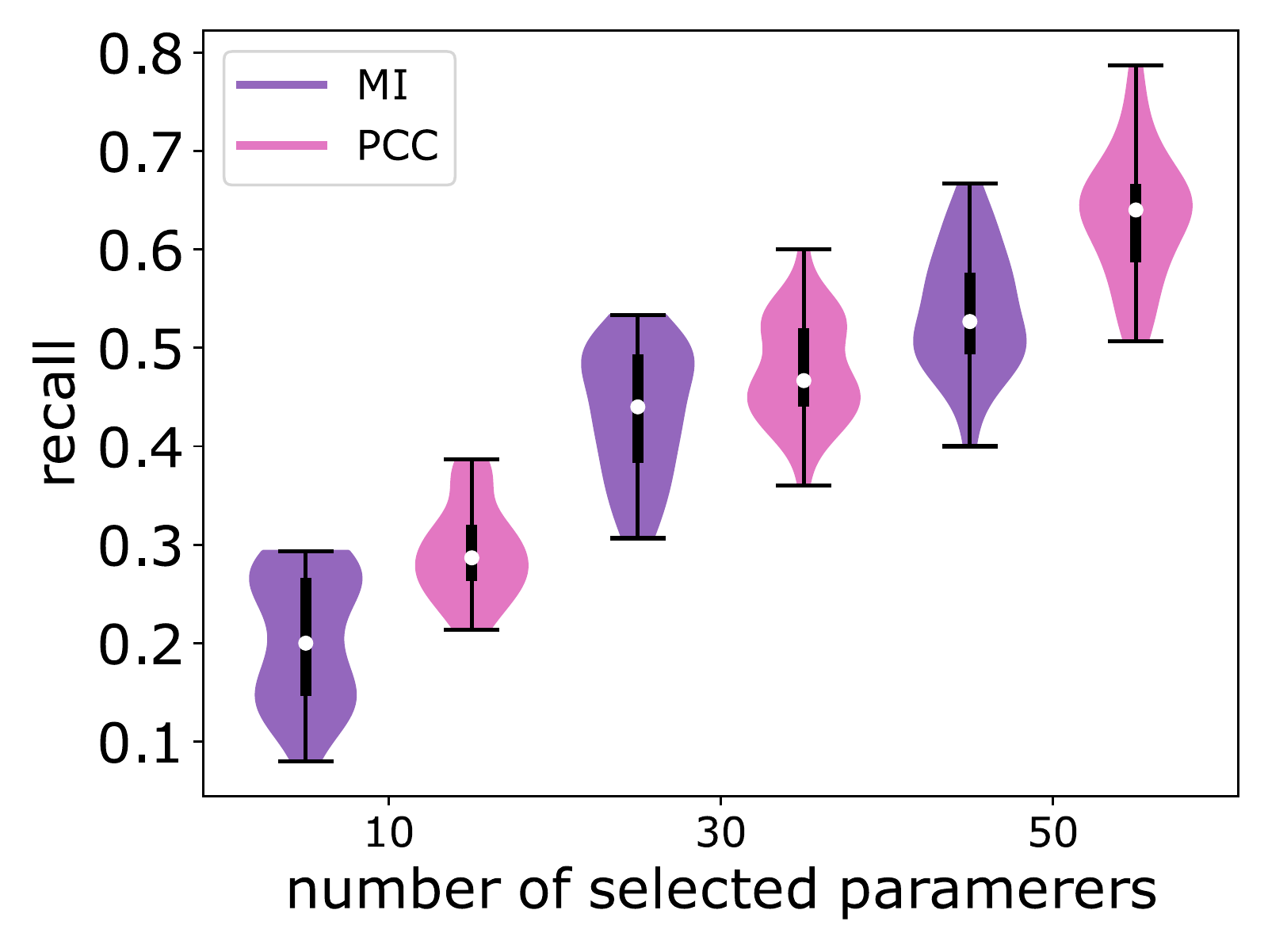}
     \end{subfigure}
        \caption{Precision and recall of effective parameter estimation}
        \label{fig:param_precision_recall}
    \end{minipage}
\end{figure*}

\begin{figure*}[t]
     \centering
     \begin{subfigure}[t]{0.31\textwidth}
         \centering
         \includegraphics[width=\textwidth]{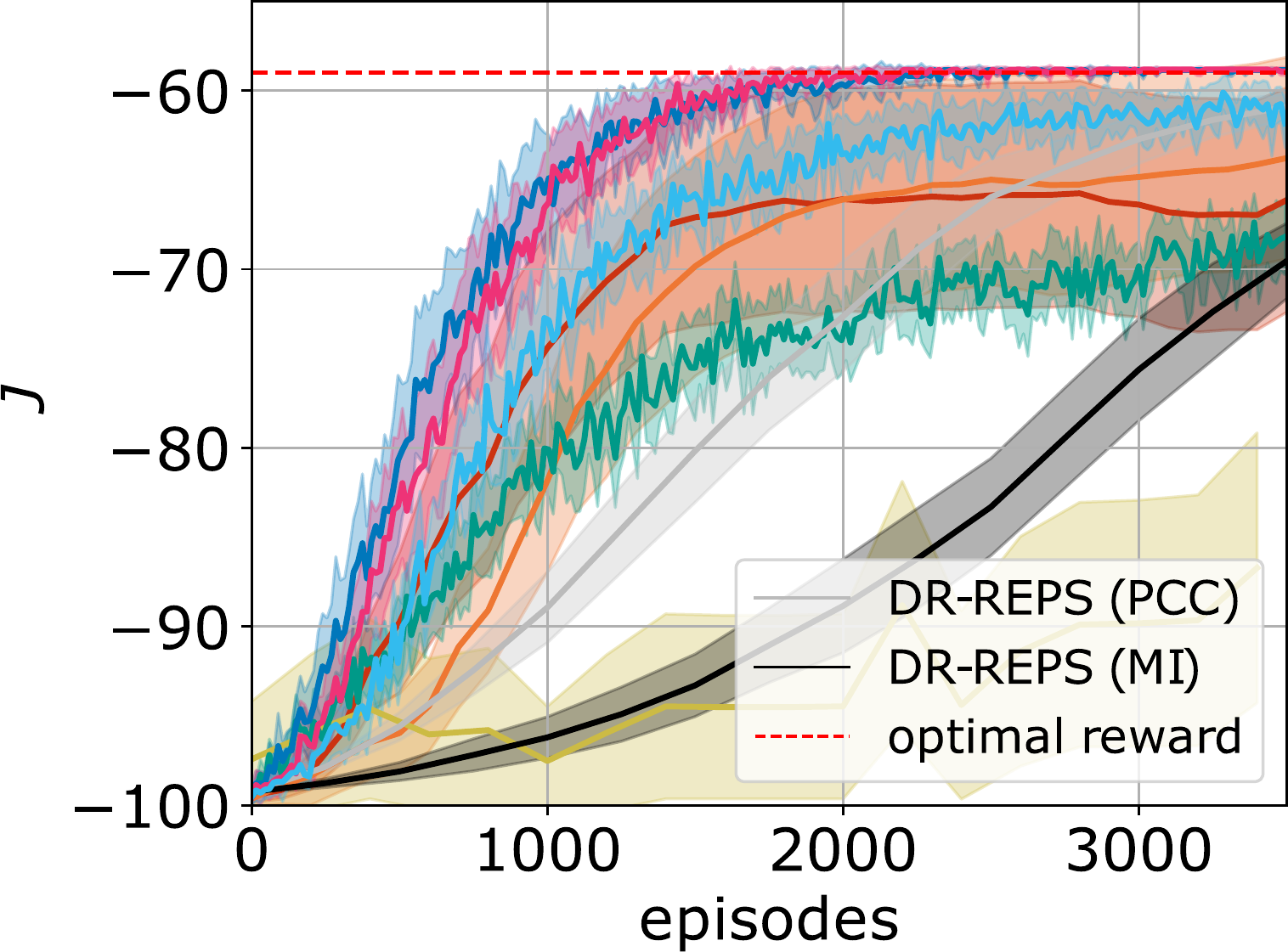}
         \caption{ShipSteering}
         \label{fig:ship_full}
     \end{subfigure}
     \hfill
     \begin{subfigure}[t]{0.31\textwidth}
         \centering
         \includegraphics[width=\textwidth]{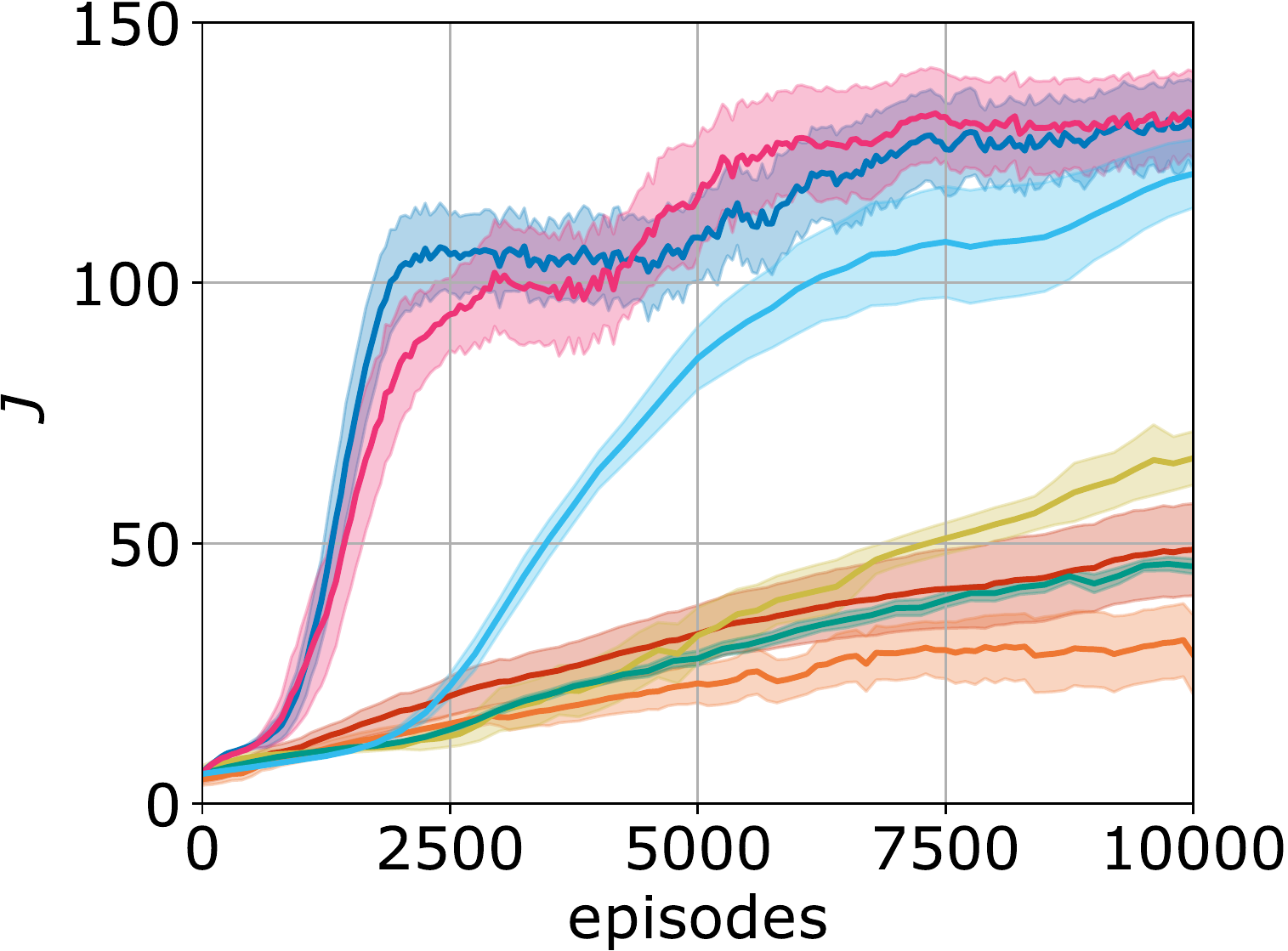}
         \caption{AirHockey}
         \label{fig:hockey_full}
     \end{subfigure}
     \hfill
     \begin{subfigure}[t]{0.31\textwidth}
         \centering
         \includegraphics[width=\textwidth]{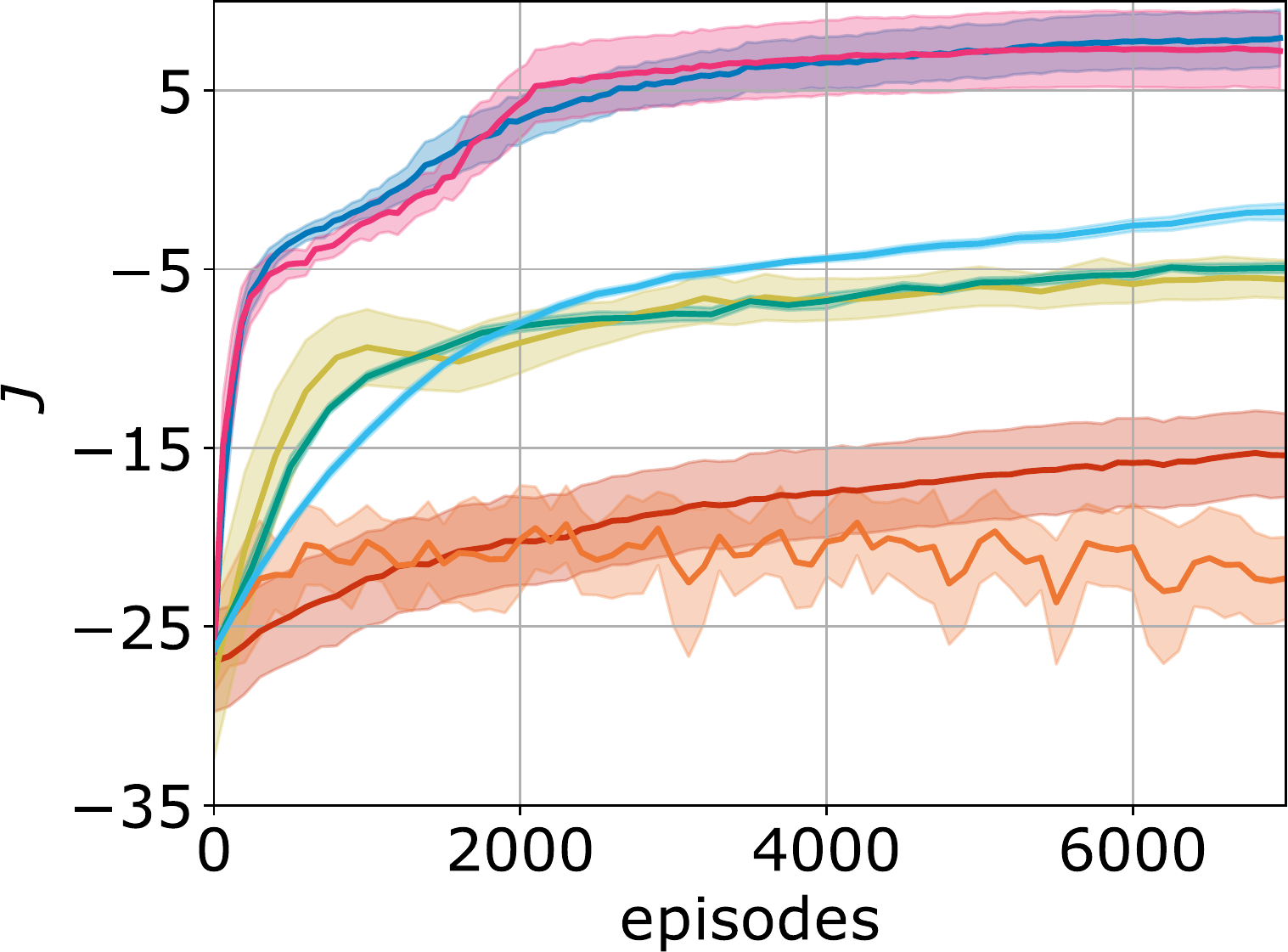}
         \caption{BallStopping}
         \label{fig:ball_full}
     \end{subfigure}
     \begin{subfigure}[b]{\textwidth}
         \vspace{1em}
         \centering
         \includegraphics[width=0.9\textwidth]{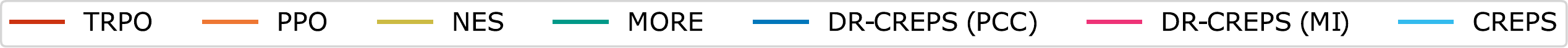}
     \end{subfigure}
    \caption{Experimental results on the simulated environments. \gls{drcreps} learns faster and converges to better optima than vanilla \gls{creps} and \gls{more}.}
    \label{fig:results_ship_hockey_ball}
\end{figure*}

We demonstrate the learning performance in the \gls{lqr} environment with full covariance matrix. When using this covariance parametrization, more episodes are required to maintain the positive definiteness during the update. We, therefore, also include the number of episodes per policy update in the parameter sweep for each algorithm.

In Fig.~\ref{fig:lqr_full}, we show the best learning performance of each algorithm. \gls{more} requires 250 episodes per policy update, \gls{creps} requires 150 episodes, \gls{trpo} and \gls{ppo} requires 100 episodes each, \gls{nes} requires 200 episodes, and \gls{drcreps} requires only 50 episodes. Consequently, \gls{drcreps} is able to perform more policy updates than \gls{more} and \gls{creps}, resulting in better learning performance. The \gls{gdr} combined with \gls{pe} can further exploit the additional updates by focusing the exploration on the effective parameters. The combination of both techniques leads to further improvements in the learning performance. For a more detailed ablation on the relationship between those methods please refer to Fig.~\ref{fig:all_ablation_random} of the Supplementary Materials.

\subsection{Comparison between MI and PCC}
We empirically evaluate the effect of \gls{mi} and \gls{pearson} as the correlation measures for selecting effective parameters in \gls{drcreps} on all environments. Both experience similar learning and achieve similar performance, as visualized in Fig.~\ref{fig:lqr_full} and Fig.~\ref{fig:results_ship_hockey_ball}.
To further investigate the performance, we again focus on the modified \gls{lqr} environment where it is easy to impose a set of effective parameters. In this modified \gls{lqr} environment, 3 parameters out of 100 are effective parameters. We estimate \gls{pearson} from Eq.~\eqref{eq:pearsonr} and the \gls{mi} with scikit-learn's mutual information regressor~\citep{scikit-learn}. We compute the precision and recall w.r.t. the identified effective parameters and show our results for different numbers of selected parameters $\{10, 30, 50\}$ in Fig.~\ref{fig:param_precision_recall}. \gls{pearson} thereby outperforms \gls{mi} by a short margin on all tested configurations which we accredit to the easier estimation of \gls{pearson} which does not require density estimation of the distributions over the parameters and the return. 
Computing the \gls{pearson} with fewer samples is therefore expected to achieve a better estimation than \gls{mi}.
Besides reducing the number of samples required for each update, our objective is also to reduce the dimensionality by assuming a small number of effective parameters. This lets us tend towards selecting less parameters, i.e., $20-50\%$ of the total parameters. These propensities explain why \gls{pearson} and \gls{mi} have similar learning performances in all experiments with \gls{pearson} being slightly superior.  We provide a more elaborate ablation study of the correlation measure as a comparison to a random selection of the parameters in Sec.~\ref{sec:corelation_comparison} of the Supplementary Materials. 
In most environments, the number of ground truth effective parameters is generally not known. It is possible that the number of selected effective parameters is much less than the real number of parameters. Misidentified parameters, however, can prevent overconfident sampling and overly restrictive policy updates, which may lead to premature convergence.

\subsection{Learning in Simulated Robotics Environments}
The simulated robotic environments are not only more difficult to solve than the \gls{lqr} but at the same time enable to validate the existence of effective parameters in real-world applications.
In all environments, \gls{gdr} allows updating the policy with fewer episodes than the original \gls{creps}. Together with \gls{pe} this results in \gls{drcreps} learning a good policy after only a fraction of the learning epochs.

The \textit{ShipSteering} environment is characterized by a high dimensional policy, i.e., 450 parameters. Learning a full covariance matrix while keeping the positive definiteness using \gls{reps} would require an unmanageable number of episodes, and it is therefore not feasible. However, if we combine \gls{reps} with \gls{gdr}, we can run the algorithm with only 250 episodes per update. We call the combination of these two approaches \gls{drreps}. The results in Fig.~\ref{fig:ship_full} show that \gls{drreps} learns a policy on par with \gls{creps} and \gls{more}, while the \gls{drcreps} algorithm outperforms all other methods.

In \textit{AirHockey}, \gls{drcreps} learns to score a goal after approx. $2500$ episodes indicated by the jump in the total return. Subsequently, it improves its reliability to shoot a goal in the subsequent episodes. \gls{more}, on the other hand, does not show the characteristic jump which is an indicator of the policy not scoring a goal. Here, small differences in the collision between two rounded-shaped objects, i.e., the puck and the end effector, result in drastically different trajectories. As a result, the cumulative return has a high variance shown in the learning curve.

\gls{drcreps} can intercept the ball after only approx. $500$ episodes in the \textit{BallStopping} environment, and the final policy learns to reliably stop it. Observing the learned policies, the advantage of our approach is clear: \gls{creps} moves the arm towards the ball but fails to reliably stops it from rolling off the table while \gls{more} completely fails to move the robotic arm towards the ball or moving the arm too rapidly, propelling the ball back of the table. These behaviors are a result of exploration in the ineffective parameters which cause the arm to act almost randomly and either move it away from the table or let it experience jerky movements. Instead, \gls{drcreps} reduces the exploration in the ineffective parameters thus, it almost completely prevents those faulty movements leading to a smoother behavior that reliably stops the ball. The \gls{gdr} then further speeds up the learning by only updating the effective parameters, i.e., the ones required to move the end effector in front of the ball.

To show the relevance of our approach, we also include \gls{drl} baselines in our evaluation, which use Deep Neural Networks as policies.
These approaches show good learning performances on the \textit{LQR} and the \textit{ShipSteering} tasks, while they fail on more difficult ones like \textit{AirHockey} and \textit{BallStopping}. 
Indeed, in robotic tasks, a structured policy rather than a neural network can significantly reduce the task complexity, and consequently, reduce the number of episodes required to solve the task.
A similar observation holds for the \gls{nes} algorithm, which requires even more episodes to maintain a population large enough to perform meaningful updates.
Our approach outperforms gradient-based and random search methods by combining more efficient sampling techniques and policy update methods with structured policies. Please refer to Sec.~\ref{sec:additional_baselines} in the Supplementary Materials for additional comparison with other baselines.

\section{CONCLUSIONS}
In this paper, we propose \acrlong{gdr}, an approach to dimensionality reduction for policy search algorithms in \gls{bbo} that enables more efficient learning of policies parameterized by full covariance matrices. We utilize the \acrlong{pearson} and the \acrlong{mi} to estimate effective parameters that contribute most to the total return. By updating only those, we reduce the number of samples required for each update allowing our algorithms to boost the learning performance. Furthermore, we present \acrlong{pe} which focuses the exploration on the effective parameters and thereby reduces the exploration space. We empirically demonstrate both methods on \gls{drcreps}, an adaptation of \gls{creps}, in four different environments including simulated robotics. \gls{drcreps} outperforms recent approaches to dimensionality reduction in \gls{bbo}. 
However, our experimental evaluation focuses mainly on robotic tasks with structured policies. Further investigation is needed to assess the scalability of our approach to Deep Network policies and to see how our method performs in comparison with more recent evolutionary strategy approaches. Anyway, structured policies are currently the state-of-the-art control method for many robotics applications, and our approach is a clean, simple, and intuitive way to improve learning performance in these scenarios.
Future work will focus on bringing the proposed methods to real-world robots and evaluating different parameter effectiveness metrics. 

\subsubsection*{Acknowledgements}
The support provided by China Scholarship Council (No. 201908080039) is acknowledged.

\bibliography{bibliography}

%%%%%%%%%%%%%%%%%%%%%%%%%%%%%%%%%%%
%%%%%% SUPPLEMENT (OPTIONAL) %%%%%%
%%%%%%%%%%%%%%%%%%%%%%%%%%%%%%%%%%%

\clearpage
\appendix

\thispagestyle{empty}

\onecolumn \makesupplementtitle

\section{ENVIRONMENT DESCRIPTIONS}
\label{sec:environment_descriptions}
In the following, we describe the environments used for testing the algorithms. If not further specified, we use the default parameters of MushroomRL~\citep{mushroom_rl}. For implementation details please visit \href{https://git.ias.informatik.tu-darmstadt.de/ias\_code/aistats2022/dr-creps}{git.ias.informatik.tu-darmstadt.de/ias\_code/aistats2022/dr-creps}.

\subsection{Linear Quadratic Regulator}
The \gls{lqr} is characterized by a linear system with quadratic rewards. The state transition and reward are defined as follows:
\begin{align*}
    x_{t+1}=Ax_{t} + Bu_{t} & & r_t=-(x_t^TQx_t + u_t^TRu_t)
\end{align*}
where $x_t$ is the state, $u_t$ is the control signal, and $r_t$ is the reward at time step $t$. Matrix $A$ denotes the state dynamics, $B$ the action dynamics and, $Q$ and $R$ the reward weights for state and action respectively. All four matrices are diagonal and we limit the actions and states to a maximum value of $1.$ to further simplify the task.
To simulate a real-world example where some parameters do not contribute to the objective, we randomly set elements on the diagonal of $Q$ and corresponding $B$ close to zero (1e-20). These dimensions then represent the ineffective parameters. We choose a 10-dimensional \gls{lqr} with 7 ineffective and 3 effective dimensions, i.e., we replace 7 values at the same positions on each diagonal.
Since the matrices remain non-singular, we can solve for the optimal control policy iteratively and assess the convergence of the algorithms to the optimal solution.

\subsection{ShipSteering}
In the ship steering task, the agent has to maneuver a ship through a gate placed at a fixed position~\citep{Ghavamzadeh2003}.
Initial position, orientation, and turning rate are chosen randomly at the start of each epoch and the ship has to maintain a constant speed while moving. The state is represented by the position (x and y) and the orientation. We transform those features using three rectangular tilings over the state dimension with the number of tiles being 5, 5, and 6, respectively. It is also straightforward to compute the optimal return by taking the shortest route from start to gate. We include this return in our plots to assess algorithm performance.

\subsection{AirHockey}
The air hockey environment features a planar robot arm with 3 \gls{dof} whose objective is to hit the puck coming towards it and score a goal on the opponents' side of the field.
During an episode, the end effector has to stay on the table and we terminate an episode if it leaves the table boundaries. We simulate with a horizon of 120 steps.

\subsection{BallStopping}
In this environment, a ball is placed on a table and starts rolling off of it. The goal then is to stop it from doing so using a 7 \gls{dof} robotic arm. A penalty is given when the ball drops off the table with a discount factor decreasing this penalty over time. We run an episode for a horizon of 750, which marks the time when the ball would've dropped off the table without any intervention by the robot.
The state consists of the joint position and velocity as well as the orientation and position of the ball, which results in a 21-dimensional observation space.

\section{EXPERIMENT DETAILS}
\label{sec:experimental_details}
In this section, we describe the sweep over the hyperparameters as well as the parameters selected for the plots in the paper. First, we describe the parametric policies as well as the search distributions for each environment in Tab.~\ref{tab:policy_params}. For the sweep we define $[\text{inclusive start} | \text{exclusive end} | \text{step size}]$ to denote the range of parameters explored. Besides Tab.~\ref{tab:hyper_lqr_diag} and Tab.~\ref{tab:hyper_precision_recall} which represent diagonal covariance cases, all rest tables in this section represent the full covariance case.

For the diagonal covariance case, we use the same number of episodes for all algorithms, since we do not apply \gls{gdr} yet. For full covariance matrices, we sweep multiples of 5 (starting from 15) and select the lowest possible number of episodes per fit that leads to a satisfaction of the constraints and does not violate the positive definiteness of the covariance matrix.
In general, we search algorithm-specific parameters ($\beta,\epsilon,\kappa$) on the vanilla versions of \gls{reps} and \gls{creps}, and use those as fixed parameters for the sweep on the remaining hyperparameters of \gls{drreps} and \gls{drcreps}. For assessing \gls{pe}, we keep the hyperparameter values from the corresponding \gls{drcreps} and just remove the exploration modification.

The standard algorithmic implementation of \gls{creps} does not always fulfill the positive definiteness of the covariance matrix, particularly for high values of the KL bound $\epsilon$. This happens due to an insufficient number of samples w.r.t. to a high dimensional covariance matrix, leading to computational errors. If this problem appears, we check the corresponding parameter values for \gls{drcreps}. In the case of \textit{ShipSteering}, \gls{drcreps} allows selecting a higher $\epsilon$, without violating the positive definiteness, than the one used by \gls{creps}.
\vspace{3cm}
\begin{table}[h]
  \centering
  \caption{Search distribution and corresponding parametric policy for each environment.}
    \begin{tabular}{lcccc}
    \toprule
          & LQR   & ShipSteering & AirHockey & BallStopping \\
    \midrule
    parametric policy & linear regressor & linear regressor & \acrshort{promp} & \acrshort{promp} \\
    bases      &   -    &     -  & radial basis function & radial basis function \\
    bases/dimension      &   -    &    -   & 30 & 20 \\
    basis width      &    -   &    -   & 0.001 & 0.001 \\
    parameters & 100   & 450   & 90    & 140 \\
    \midrule
    search distribution & $\mathcal{N}(\bm{\mu},\bm{\Sigma})$ & $\mathcal{N}(\bm{\mu},\bm{\Sigma})$ & $\mathcal{N}(\bm{\mu},\bm{\Sigma})$ & $\mathcal{N}(\bm{\mu},\bm{\Sigma})$ \\
    $\bm{\mu}_\text{init}$ & $\bm{0}$ & $\bm{0}$ & $\bm{0}$ & $\bm{0}$ \\
    $\bm{\Sigma}_\text{init}=diag(\bm{\sigma}_0, ...,\bm{\sigma}_n)$ & $\bm{\sigma}=0.3$ & $\bm{\sigma}=0.07$ & $\bm{\sigma}=1.0$ & $\bm{\sigma}=1.0$ \\
    \bottomrule
    \end{tabular}%
  \label{tab:policy_params}%
\end{table}%

\begin{table}[b]
  \centering
  \caption{Hyperparameters for Fig.~\ref{fig:lqr_diag}, LQR - Diagonal Covariance Matrix \\
  }
    \begin{tabular}{lcccccc}
    \toprule
          & RWR   & PRO   & REPS  & REPS & CREPS & CREPS \\
          &    &    &   &  w/ PE &  &  w/PE \\
    \midrule
    \textbf{Sweep} &      &      &      &      &      &  \\
    $\beta$ & $[0.1|1.5|0.1]$ & -     & -     & -     & -     & - \\
    $\epsilon$ & -     & -     & $[0.1|1.5|0.1]$ & -     & $[1.5|3.5|0.2]$ & - \\
    $\kappa$ & -     & -     & -     & -     & $[1.|10.|1.]$ & - \\
    $m$ & -     & -     & -     & $[5|100|5]$ & -     & $[5|100|5]$ \\
    $\gamma$ & -     & -     & -     & $[0.1|1.0|0.1]$ & -     & $[0.1|1.0|0.1]$ \\
    \midrule
    \textbf{Experiment} &       &       &       &       &       &  \\
    $\beta$ & 0.2   & 0.2   & -     & -   & -     & - \\
    $\epsilon$ & -     & -     & 0.4   & 0.4     & 2.5   & 2.5 \\
    $\kappa$ & -     & -     & -     & -     & 6.0   & 6.0 \\
    DR    & no    & no    & no    & no    & no    & no \\
    PE    & no    & no    & no    & yes   & no    & yes \\
    $m$ & -     & -     & -     & 30    & -     & 30 \\
    $\gamma$ & -     & -     & -     & 0.1/0.9 & -     & 0.1 \\
    correlation measure & -     & PCC   & -     & MI    & -     & MI \\
    number of epochs & 80    & 80    & 80    & 80    & 80    & 80 \\
    episodes per fit & 25    & 25    & 25    & 25    & 25    & 25 \\
    \bottomrule
    \end{tabular}%
  \label{tab:hyper_lqr_diag}%
\end{table}%

\begin{table}[b]
  \centering
  \caption{Hyperparameters for Fig.~\ref{fig:lqr_full} and Fig.~\ref{fig:additional_baselines}, LQR - Full Covariance Matrix}
    \begin{tabular}{lcccccc}
    \toprule
          & MORE  & CREPS & DR-CREPS & DR-CREPS & DR-CREPS & DR-CREPS \\
          &   &  & (PCC) & w/o PE & (MI) &w/o PE \\
          &   &  &  & (PCC) & & (MI) \\
    \midrule
    \textbf{Sweep} &       &       &       &       &       &  \\
    $\epsilon$ & $[1.9|6.0|0.2]$ & $[1.9|6.0|0.2]$ & -     & -     & -     & - \\
    $\kappa$ & $[1.|20.|1.]$ & $[1.|20.|1.]$ & -     & -     & -     & - \\
    $m$ & -     & -     & $[5|100|5]$ & -     & $[5|100|5]$ & - \\
    $\gamma$ & -     & -     & $[0.1|1.0|0.1]$ & -     & $[0.1|1.0|0.1]$ & - \\
    \midrule
    \textbf{Experiment} &       &       &       &       &       &  \\
    $\epsilon$ & 4.7   & 4.7   & 4.7   & 4.7   & 4.7   & 4.7 \\
    $\kappa$ & 17.0  & 17.0  & 17.0  & 17.0  & 17.0  & 17.0 \\
    DR    & no    & no    & yes   & yes   & yes   & yes \\
    PE    & no    & no    & yes   & no    & yes   & no \\
    $m$ & -     & -     & 50    & 50    & 50    & 50 \\
    $\gamma$ & -     & -     & 0.1   & 0.1   & 0.1   & 0.1 \\
    correlation measure & -     & -     & PCC   & PCC   & MI    & MI \\
    number of epochs & 20    & 33    & 100   & 100   & 100   & 100 \\
    episodes per fit & 250   & 150   & 50    & 50    & 50    & 50 \\
    \bottomrule
    \end{tabular}%
  \label{tab:hyper_lqr_full_0}%
\end{table}%

\begin{table}[b]
  \centering
  \caption{Hyperparameters for Fig.~\ref{fig:lqr_full} and Fig.~\ref{fig:additional_baselines}, LQR - Full Covariance Matrix}
    \begin{tabular}{lccccc}
    \toprule
          & TRPO  & PPO   & REINFORCE & NES   & CEM \\
    \midrule
    \textbf{Sweep} &       &       &       &       &  \\
    actor lr & -& [3e-2, 3e-3, 3e-4] & -     & -     & - \\
    citic lr &  [3e-2, 3e-3, 3e-4]     & [3e-2, 3e-3, 3e-4] & -     & -     & - \\
    max KL & [1e-0, 1e-1, 1e-2] & -     & -     & -     & - \\
    optimizer step size $\epsilon$ & -     & -     & [1e-1, 1e-2, 1e-3] & -     & - \\
    optimizer lr & -     & -     & -     & [3e-1, 3e-2, 3e-3] & - \\
    population size & -     & -     & -     & [50, 100, 200, 500] & - \\
    \textbf{Experiment} & -     & -     & -     & -     & - \\
    actor lr &  -  &  3e-3 & -     & -     & - \\
    citic lr &  3e-3   &  3e-3 & -     & -     & - \\
    max KL & 1e-0  & -     & -     & -     & - \\
    optimizer step size $\epsilon$ & -     & -     & 1e-2  & -     & - \\
    optimizer lr & -     & -     & -     & 3e-2  & - \\
    rollouts & -     & -     & -     & 2     & - \\
    population size & -     & -     & -     & 100   & - \\
    elites & -     & -     & -     & -     & 25 \\
    number of epochs & 50    & 50    & 50    & 200   & 100 \\
    episodes per fit & 100   & 100   & 100   & 25    & 50 \\
    \bottomrule
    \end{tabular}%
  \label{tab:hyper_lqr_full_1}%
\end{table}%

\begin{table}[b]
  \centering
  \caption{Hyperparameters for Fig.~\ref{fig:param_precision_recall}, Precision and Recall for Effective Parameter Estimation}
    \begin{tabular}{lc}
    \toprule
          & DR-CREPS \\
    \midrule
    \textbf{Experiment} &  \\
    $\epsilon$ & 4.5 \\
    $\kappa$ & 15. \\
    DR    & no \\
    PE    & yes \\
    $m$ & 10/30/50 \\
    $\gamma$ & 0.1 \\
    correlation measure & PCC/MI \\
    number of epochs & 40 \\
    episodes per fit & 50 \\
    \bottomrule
    \end{tabular}%
  \label{tab:hyper_precision_recall}%
\end{table}%

\begin{table}[b]
  \centering
  \caption{Hyperparameters for Fig.~\ref{fig:ship_full} and Fig.~\ref{fig:additional_baselines}, ShipSteering - Full Covariance Matrix}
    \begin{tabular}{lcccccc}
    \toprule
          & DR-REPS & DR-REPS & MORE  & CREPS & DR-CREPS & DR-CREPS \\
          &(PCC) &(MI) &   &  & (PCC) & (MI) \\
  \midrule
    \textbf{Sweep} &       &       &       &       &       &  \\
    $\epsilon$ &   -    &    -   & $[2.4|6.4|1.0]$ & $[1.9|3.9|0.5]$ & $[2.4|4.4|1.0]$ & $[2.4|4.4|1.0]$ \\
    $\kappa$ &    -   &   -    &    -   & $[14|30|3]$ &  -     &  \\
    $m$ & $[100|500|50]$ & $[100|500|50]$ &  -     &    -   & $[100|500|50]$ & $[100|500|50]$ \\
    $\gamma$ & $[0.1|0.9|0.2]$ & $[0.1|0.9|0.2]$ &   -    &   -    & $[0.1|0.9|0.2]$ & $[0.1|0.9|0.2]$ \\
    \midrule
    \textbf{Experiment} &       &       &       &       &       &  \\
    $\epsilon$ & 0.5   & 0.5   & 4.4   & 2.4   & 3.4   & 3.4 \\
    $\kappa$ &    -   &   -    & 20.   & 20.   & 20.   & 20. \\
    DR    & yes    & yes    & no    & no    & yes   & yes \\
    PE    & yes    & yes    & no    & no    & yes   & yes \\
    $m$ & 100   & 100   & -     & -     & 200   & 200 \\
    $\gamma$ & 0.1   & 0.1     & -     &  -     & 0.1   & 0.1 \\
    correlation measure & PCC   & MI    & -     & -     & PCC   & MI \\
    number of epochs & 14    & 14    & 233   & 233   & 233   & 233 \\
    episodes per fit & 250   & 250   & 15    & 15    & 15    & 15 \\
    \bottomrule
    \end{tabular}%
  \label{tab:hyper_ship_steering_0}%
\end{table}%

\begin{table}[b]
  \centering
  \caption{Hyperparameters fo Fig.~\ref{fig:ship_full} and Fig.~\ref{fig:additional_baselines}, ShipSteering - Full Covariance Matrix}
    \begin{tabular}{lccccc}
    \toprule
          & TRPO  & PPO   & REINFORCE & NES   & CEM \\
    \midrule
    \textbf{Sweep} &       &       &       &       &  \\
    actor lr & -  & [3e-2, 3e-3, 3e-4] & -     & -     & - \\
    citic lr &  [3e-2, 3e-3, 3e-4]   & [3e-2, 3e-3, 3e-4] & -     & -     & - \\
    max KL & [1e-0, 1e-1, 1e-2] & -     & -     & -     & - \\
    optimizer step size $\epsilon$ & -     & -     & [1e-1, 1e-2, 1e-3] & -     & - \\
    optimizer lr & -     & -     & -     & [3e-1, 3e-2, 3e-3] & - \\
    population size & -     & -     & -     & [50, 100, 200, 500] & - \\
    \midrule
    \textbf{Experiment} & -     & -     & -     & -     & - \\
    actor lr & - & 3e-4  & -     & -     & - \\
    citic lr &   3e-2    & 3e-4  & -     & -     & - \\
    max KL & 1e-2  & -     & -     & -     & - \\
    optimizer step size $\epsilon$ & -     & -     & 1e-3  & -     & - \\
    optimizer lr & -     & -     & -     & 3e-2  & - \\
    rollouts & -     & -     & -     & 2     & - \\
    population size & -     & -     & -     & 100   & - \\
    elites & -     & -     & -     & -     & 25 \\
    number of epochs & 35    & 35    & 35    & 200   & 70 \\
    episodes per fit & 100   & 100   & 100   & 17    & 50 \\
    \bottomrule
    \end{tabular}%
  \label{tab:hyper_ship_steering_1}%
\end{table}%

\begin{table}[b]
  \centering
  \caption{Hyperparameters for Fig.~\ref{fig:hockey_full} and Fig.~\ref{fig:additional_baselines}, AirHockey - Full Covariance Matrix}
    \begin{tabular}{lcccc}
    \toprule
          & MORE  & CREPS & DR-CREPS & DR-CREPS \\
          &   &  & (PCC) &  (MI) \\
  \midrule
    \textbf{Sweep} &       &       &       &  \\
    $\epsilon$ & $[1.4|8.4|1.0]$ & $[0.4|3.4|0.5]$ &    -   &  -\\
    $\kappa$ & $[6.|20.|2.]$ & $[6.|20.|2.]$ &    -   &  -\\
    $m$ &  -     &     -  & $[10|90|10]$ & $[10|90|10]$ \\
    $\gamma$ &   -    &  -     & $[0.1|0.9|0.2]$ & $[0.1|0.9|0.2]$ \\
    \midrule
    \textbf{Experiment} &       &       &       &  \\
    $\epsilon$ & 2.4   & 2.0   & 2.0   & 2.0 \\
    $\kappa$ & 12.   & 12.   & 12.   & 12. \\
    DR    & no    & no    & yes   & yes \\
    PE    & no    & no    & yes   & yes \\
    $m$ &   -    &    -   & 30    & 30 \\
    $\gamma$ &    -   &    -   & 0.5   & 0.5 \\
    correlation measure &   -    &  - &   PCC  &  MI \\
    number of epochs & 40    & 40    & 200   & 200 \\
    episodes per fit & 250   & 250   & 50    & 50 \\
    \bottomrule
    \end{tabular}%
  \label{tab:hyper_air_hockey_0}%
\end{table}%

\begin{table}[b]
  \centering
  \caption{Hyperparameters for Fig.~\ref{fig:hockey_full} and Fig.~\ref{fig:additional_baselines}, AirHockey - Full Covariance Matrix}
    \begin{tabular}{lccccc}
    \toprule
          & TRPO  & PPO   & REINFORCE & NES   & CEM \\
    \midrule
    \textbf{Sweep} &       &       &       &       &  \\
    actor lr & -  & [3e-2, 3e-3, 3e-4] & -     & -     & - \\
    citic lr &  [3e-2, 3e-3, 3e-4]   & [3e-2, 3e-3, 3e-4] & -     & -     & - \\
    max KL & [1e-0, 1e-1, 1e-2] & -     & -     & -     & - \\
    optimizer step size $\epsilon$ & -     & -     & [1e-1, 1e-2, 1e-3] & -     & - \\
    optimizer lr & -     & -     & -     & [3e-1, 3e-2, 3e-3] & - \\
    population size & -     & -     & -     & [50, 100, 200, 500] & - \\
    \textbf{Experiment} & -     & -     & -     & -     & - \\
    actor lr & -  & 3e-3  & -     & -     & - \\
    citic lr &    3e-3  & 3e-3  & -     & -     & - \\
    max KL & 1e-1  & -     & -     & -     & - \\
    optimizer step size $\epsilon$ & -     & -     & 1e-1  & -     & - \\
    optimizer lr & -     & -     & -     & 3e-1  & - \\
    rollouts & -     & -     & -     & 2     & - \\
    population size & -     & -     & -     & 100   & - \\
    elites & -     & -     & -     & -     & 25 \\
    number of epochs & 100   & 100   & 100   & 200   & 200 \\
    episodes per fit & 100   & 100   & 100   & 50    & 50 \\
    \bottomrule
    \end{tabular}%
  \label{tab:hyper_air_hockey_1}%
\end{table}%

\begin{table}[b]
  \centering
  \caption{Hyperparameters for Fig.~\ref{fig:ball_full} and Fig.~\ref{fig:additional_baselines}, BallStopping - Full Covariance Matrix}
    \begin{tabular}{lcccc}
    \toprule
          & MORE  & CREPS & DR-CREPS & DR-CREPS \\
          &   &  & (PCC) & (MI) \\
  \midrule
    \textbf{Sweep} &       &       &       &  \\
    $\epsilon$ &   -    & $[1.5|6.0|0.5]$ &   -    & - \\
    $\kappa$ &   -    & $[5|30|5]$ &   -    &  -\\
    $m$ &   -    &    -   & $[15|140|15]$ & $[15|140|15]$ \\
    $\gamma$ &    -   &   -    & $[0.1|1.0|0.5]$ & $[0.1|1.0|0.5]$ \\
    \midrule
    \textbf{Experiment} &       &       &       &  \\
    $\epsilon$ & 4.5   & 4.5   & 4.5   & 4.5 \\
    $\kappa$ & 20.   & 20.   & 20.   & 20. \\
    DR    &    no   &   no    &     yes  & yes \\
    PE    &      no & no       &     yes  & yes \\
    $m$ &  -     &    -   & 30    & 30 \\
    $\gamma$ &     -  &  -     & 0.5   & 0.5 \\
    correlation measure & -      &    -   & PCC   & MI \\
    number of epochs & 28    & 28    & 116   & 116 \\
    episodes per fit & 250   & 250   & 60    & 60 \\
    \bottomrule
    \end{tabular}%
  \label{tab:hyper_ball_stopping_0}%
\end{table}%

\begin{table}[b]
  \centering
  \caption{Hyperparameters for Fig.~\ref{fig:ball_full} and Fig.~\ref{fig:additional_baselines}, BallStopping - Full Covariance Matrix}
    \begin{tabular}{lccccc}
    \toprule
          & TRPO  & PPO   & REINFORCE & NES   & CEM \\
    \midrule
    \textbf{Sweep} &       &       &       &       &  \\
    actor lr & -     & [3e-2, 3e-3, 3e-4] & -     & -     & - \\
    citic lr & [3e-2, 3e-3, 3e-4] & [3e-2, 3e-3, 3e-4] & -     & -     & - \\
    max KL & [1e-0, 1e-1, 1e-2] & -     & -     & -     & - \\
    optimizer step size $\epsilon$ & -     & -     & [1e-1, 1e-2, 1e-3] & -     & - \\
    optimizer lr & -     & -     & -     & [3e-1, 3e-2, 3e-3] & - \\
    population size & -     & -     & -     & [50, 100, 200, 500] & - \\
    \textbf{Experiment} & -     & -     & -     & -     & - \\
    actor lr & -     & 3e-3  & -     & -     & - \\
    citic lr & 3e-2  & 3e-3  & -     & -     & - \\
    max KL & 1e-0  & -     & -     & -     & - \\
    optimizer step size $\epsilon$ & -     & -     & 1e-3  & -     & - \\
    optimizer lr & -     & -     & -     & 3e-1  & - \\
    rollouts & -     & -     & -     & 2     & - \\
    population size & -     & -     & -     & 100   & - \\
    elites & -     & -     & -     & -     & 25 \\
    number of epochs & 100   & 100   & 100   & 200   & 140 \\
    episodes per fit & 70    & 70    & 70    & 35    & 50 \\
    \bottomrule
    \end{tabular}%
  \label{tab:hyper_ball_stopping_1}%
\end{table}%

\clearpage

\section{ADDITIONAL BASELINES}
 \label{sec:additional_baselines}
\begin{figure}[h]
     \centering
     \begin{subfigure}[b]{0.48\textwidth}
         \centering
         \includegraphics[width=\textwidth]{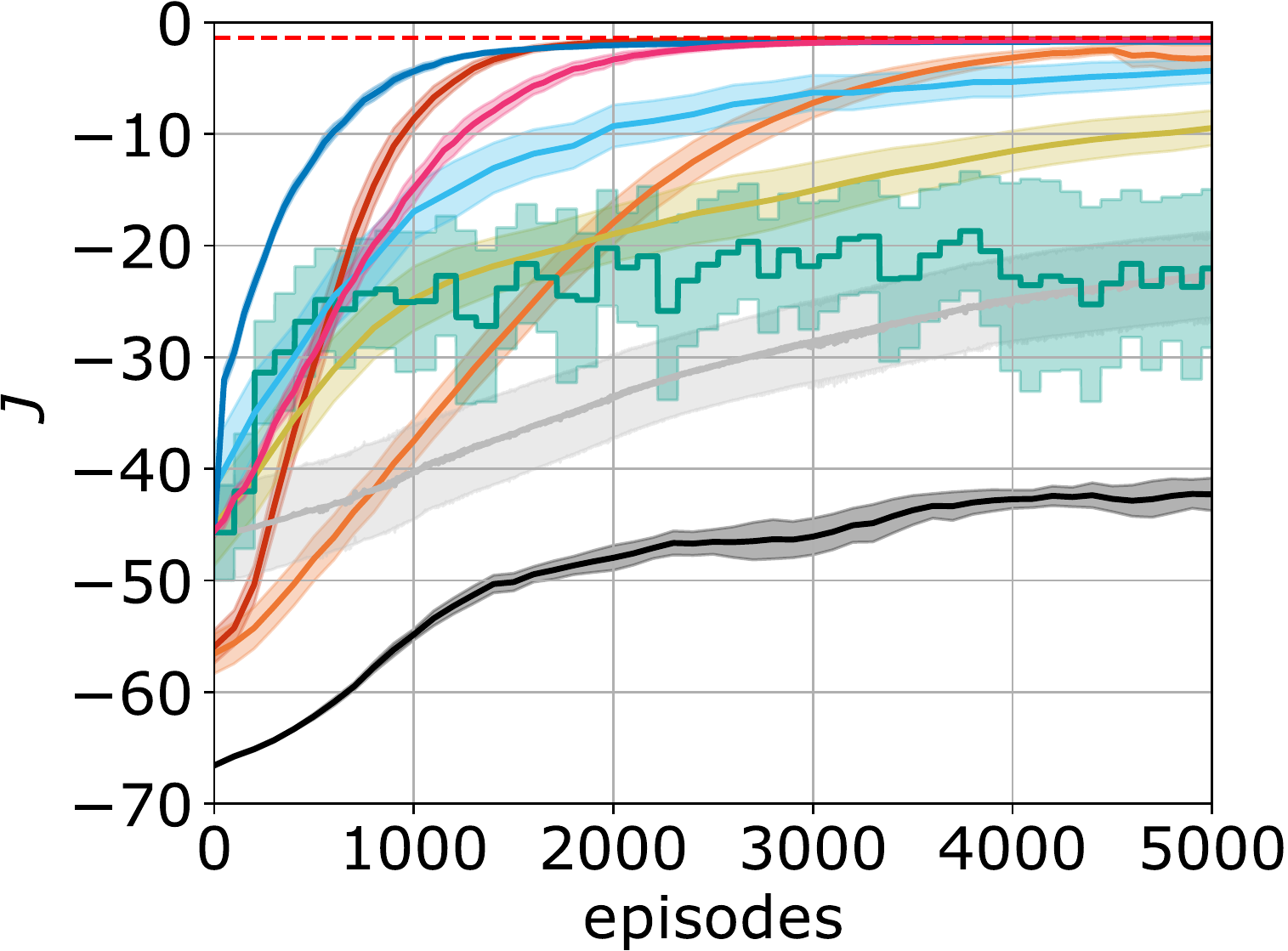}
         \caption{\gls{lqr}}
     \end{subfigure}
     \hfill
     \begin{subfigure}[b]{0.48\textwidth}
         \centering
         \includegraphics[width=\textwidth]{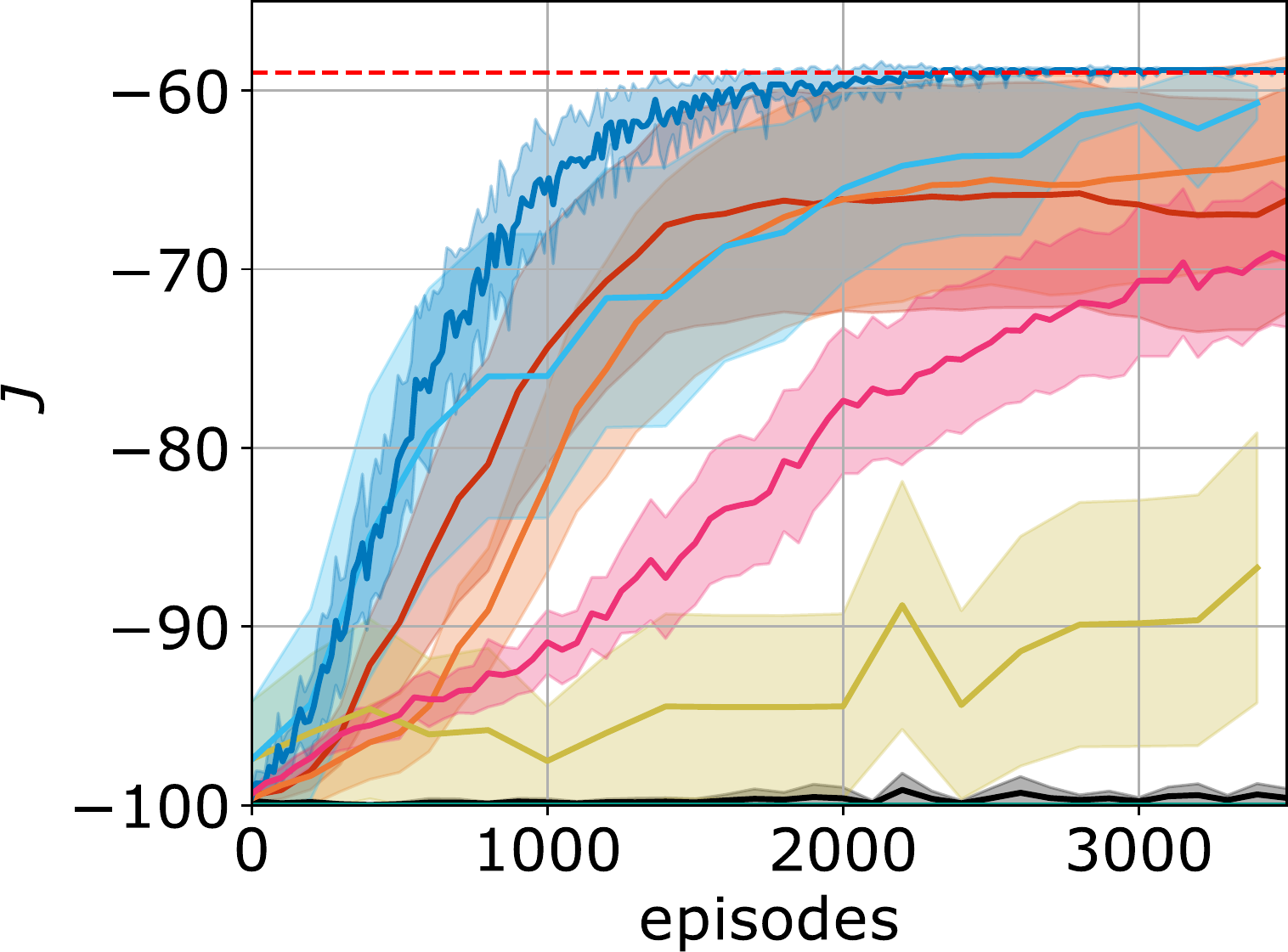}
         \caption{ShipSteering}
     \end{subfigure}
     \\
     \vspace{+1em}
     \begin{subfigure}[b]{0.48\textwidth}
         \centering
         \includegraphics[width=\textwidth]{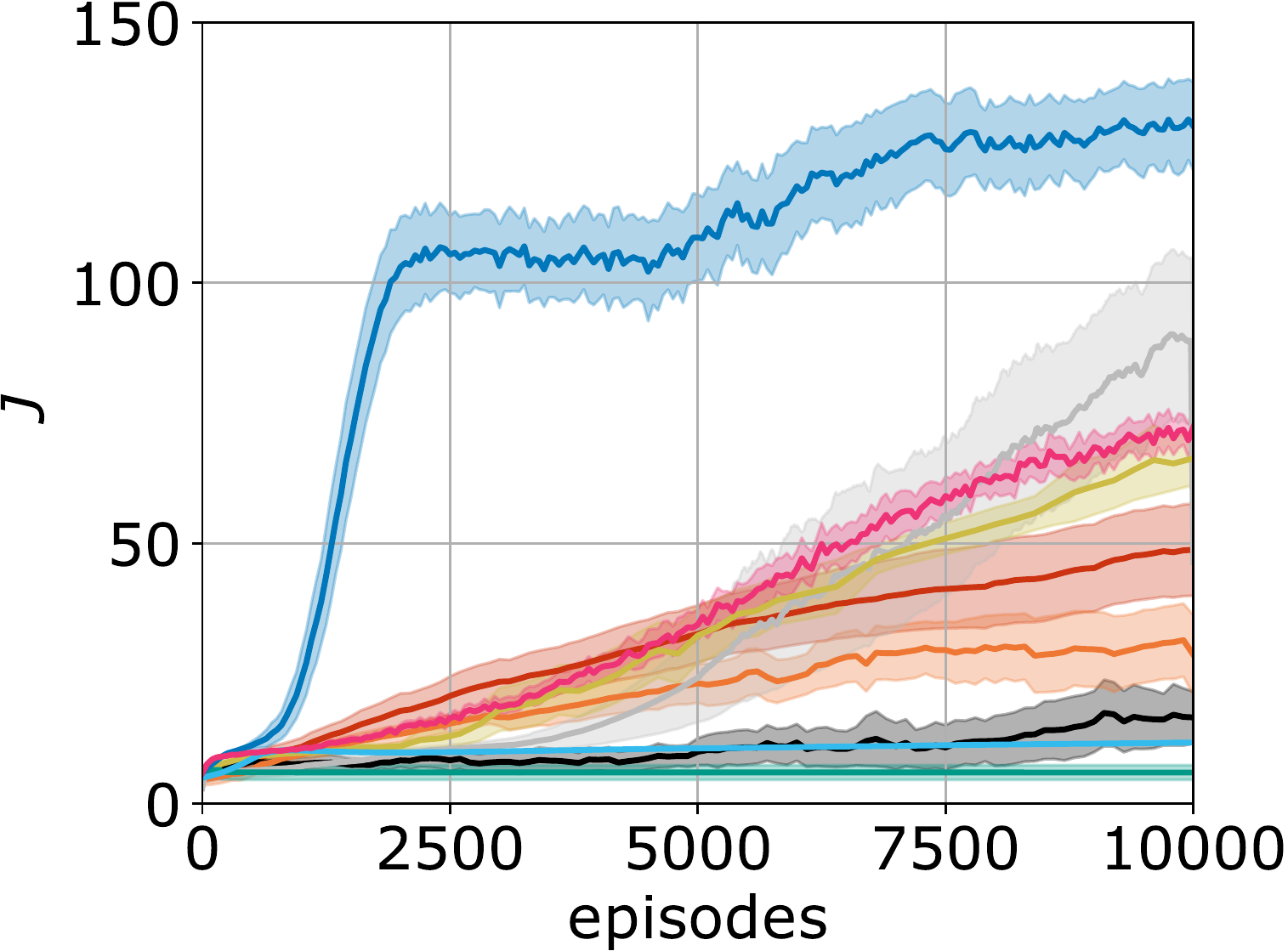}
         \caption{AirHockey}
     \end{subfigure}
     \hfill
     \begin{subfigure}[b]{0.48\textwidth}
         \centering
         \includegraphics[width=\textwidth]{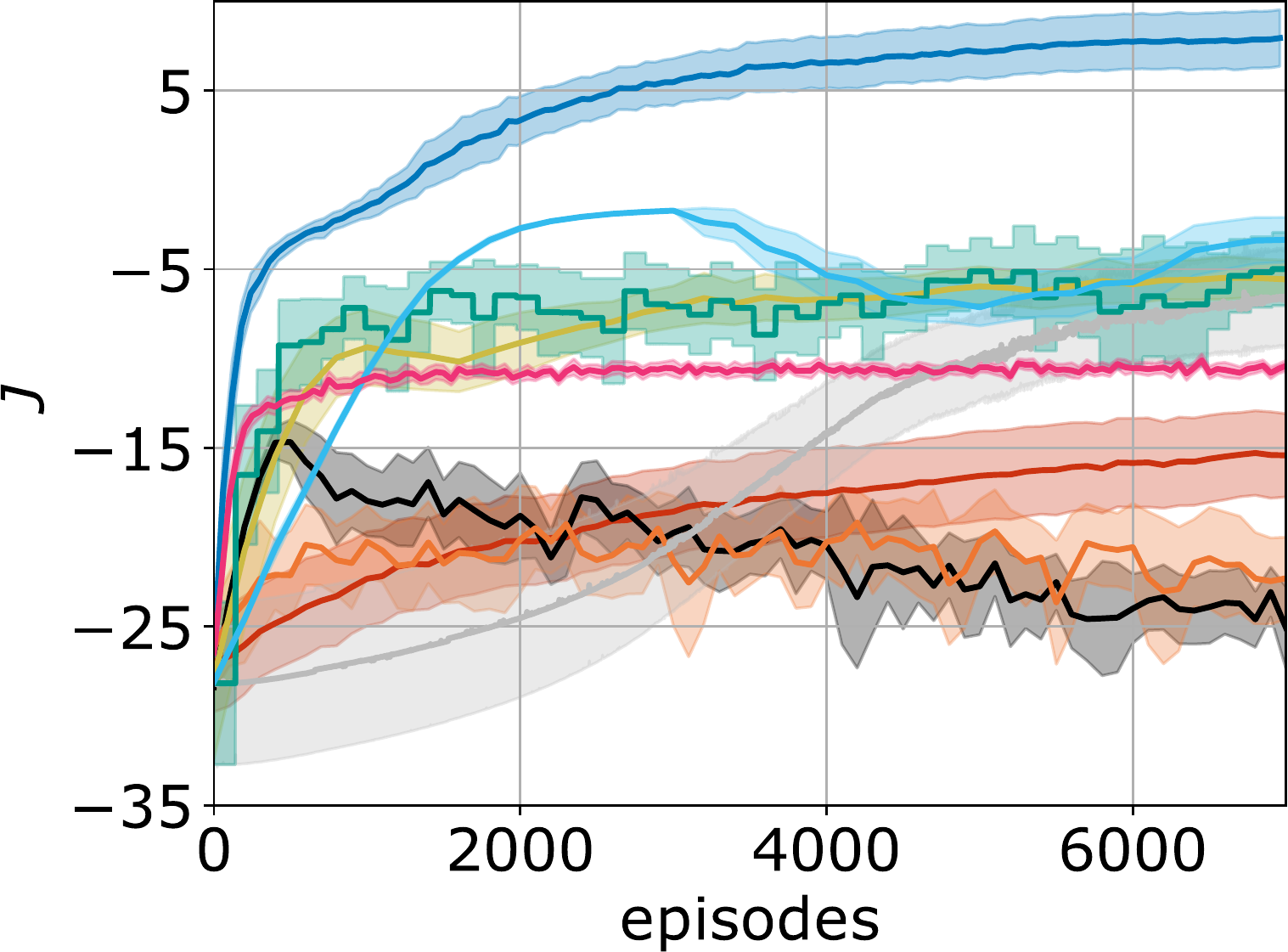}
         \caption{BallStopping}
     \end{subfigure}
    \hfill
     \begin{subfigure}[b]{\textwidth}
         \centering
         \vspace{+1em}
         \includegraphics[width=\textwidth]{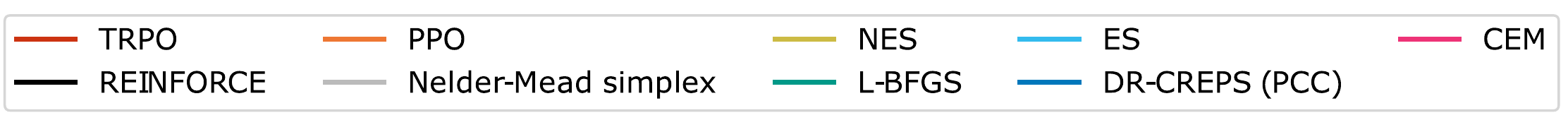}
     \end{subfigure}
    \caption{Additional baselines for all environments.}
    \label{fig:additional_baselines}
\end{figure}

We provide an additional comparison of our method to gradient-based methods (\gls{trpo}~\citep{schulman2015trust},  \gls{ppo}~\citep{schulman2017proximal}, REINFORCE \citep{Williams1992}), evolutionary strategies (\gls{es}, \gls{nes}~\citep{wierstra2014natural}), the \gls{cem}~\citep{rubinstein1999cross}, and classic optimization algorithms (\gls{nm}, \gls{lbfgs}~\citep{fletcher1987bfgs}) in Fig.~\ref{fig:additional_baselines}. Details on the hyperparameter sweep are given in Sec.~\ref{sec:experimental_details}. %Note that \gls{trpo} and \gls{ppo} fail to solve the \textit{ShipSteering} task if tiles are used as features. We therefore discard those and directly use the standardized features from the environment.
Nonetheless, our proposed approach outperforms all baselines by a significant margin with respect to the speed of learning and the final learned policy.

\section{ABLATION STUDIES}
To investigate the contributions of each of our components, we conduct an ablation study. We analyze the influence of hyperparameter $\lambda$ on the \gls{pe} and examine its interplay with the number of selected parameters. Finally, we compare the \gls{mi} and the \gls{pearson} to a random selector as means of selecting the effective parameters. If not further specified, we use the hyperparameters described in Sec.~\ref{sec:experimental_details}.

\subsection{Influence of the number of selected parameters \texorpdfstring{$m$}{Lg} and hyperparameter \texorpdfstring{$\lambda$}{Lg}}
\label{sec:influence_lambda_m}
\begin{figure}[t]
     \centering
     \begin{subfigure}[b]{0.48\textwidth}
         \centering
         \includegraphics[width=\textwidth]{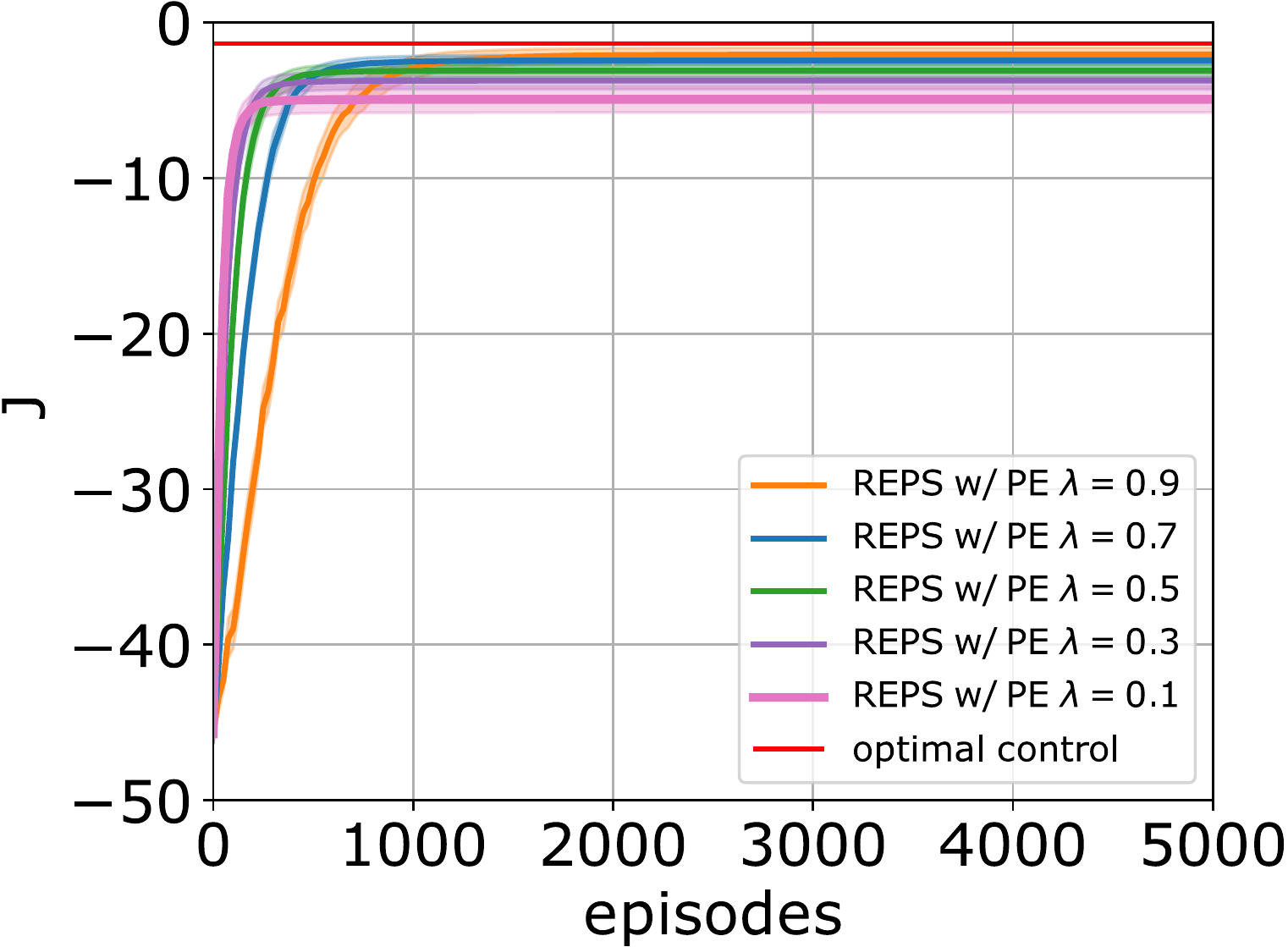}
         \caption{\gls{lqr} - DiagCov \\30 selected parameters}
         \label{fig:lqr_ablation_creps_diag}
     \end{subfigure}
     \hfill
     \begin{subfigure}[b]{0.48\textwidth}
         \centering
         \includegraphics[width=\textwidth]{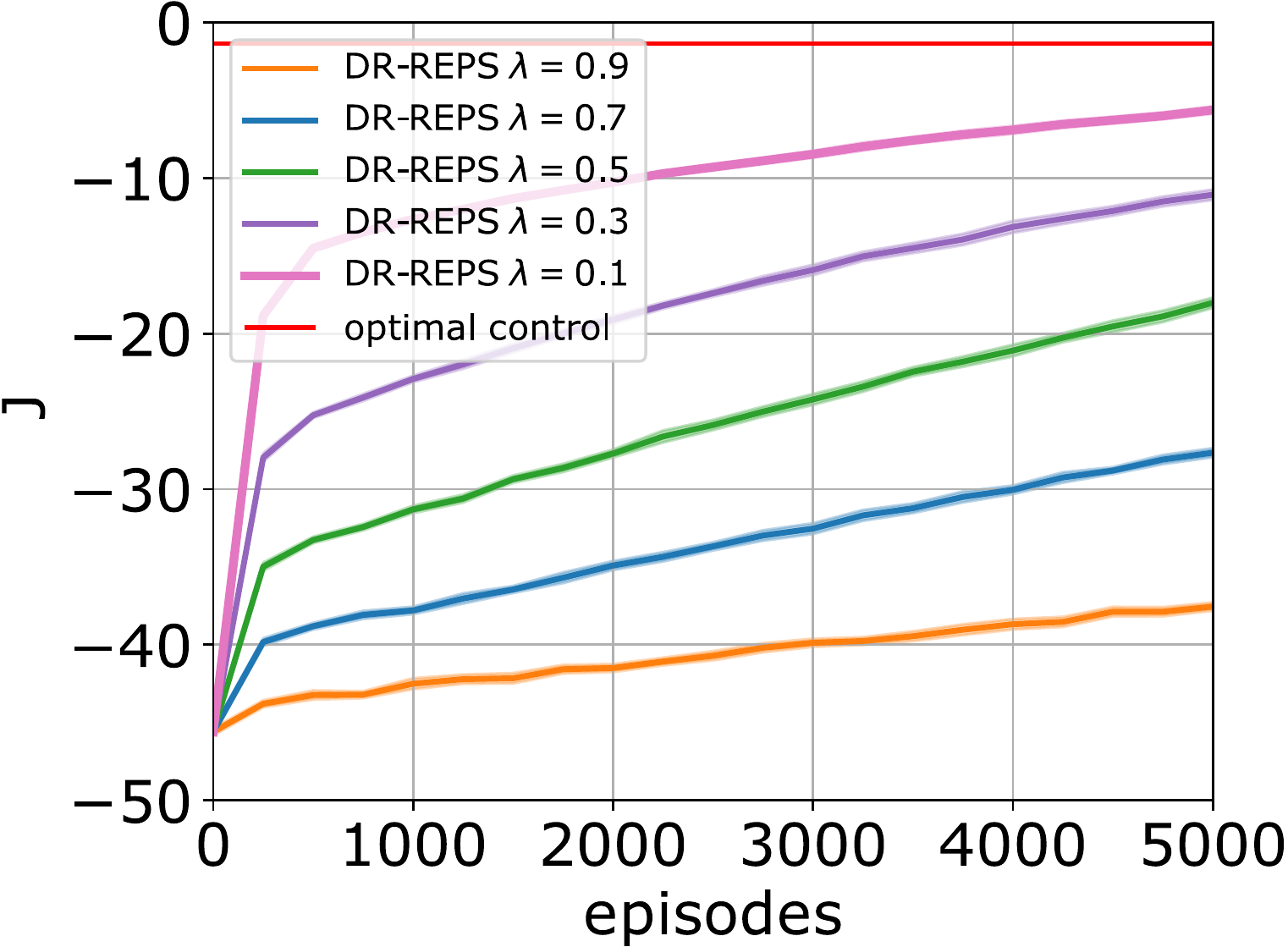}
         \caption{\gls{lqr} - FullCov \\10 selected parameters}
         \label{fig:lqr_ablation_creps_full_k_10}
     \end{subfigure}
     \\
     \vspace{+1em}
     \begin{subfigure}[b]{0.48\textwidth}
         \centering
         \includegraphics[width=\textwidth]{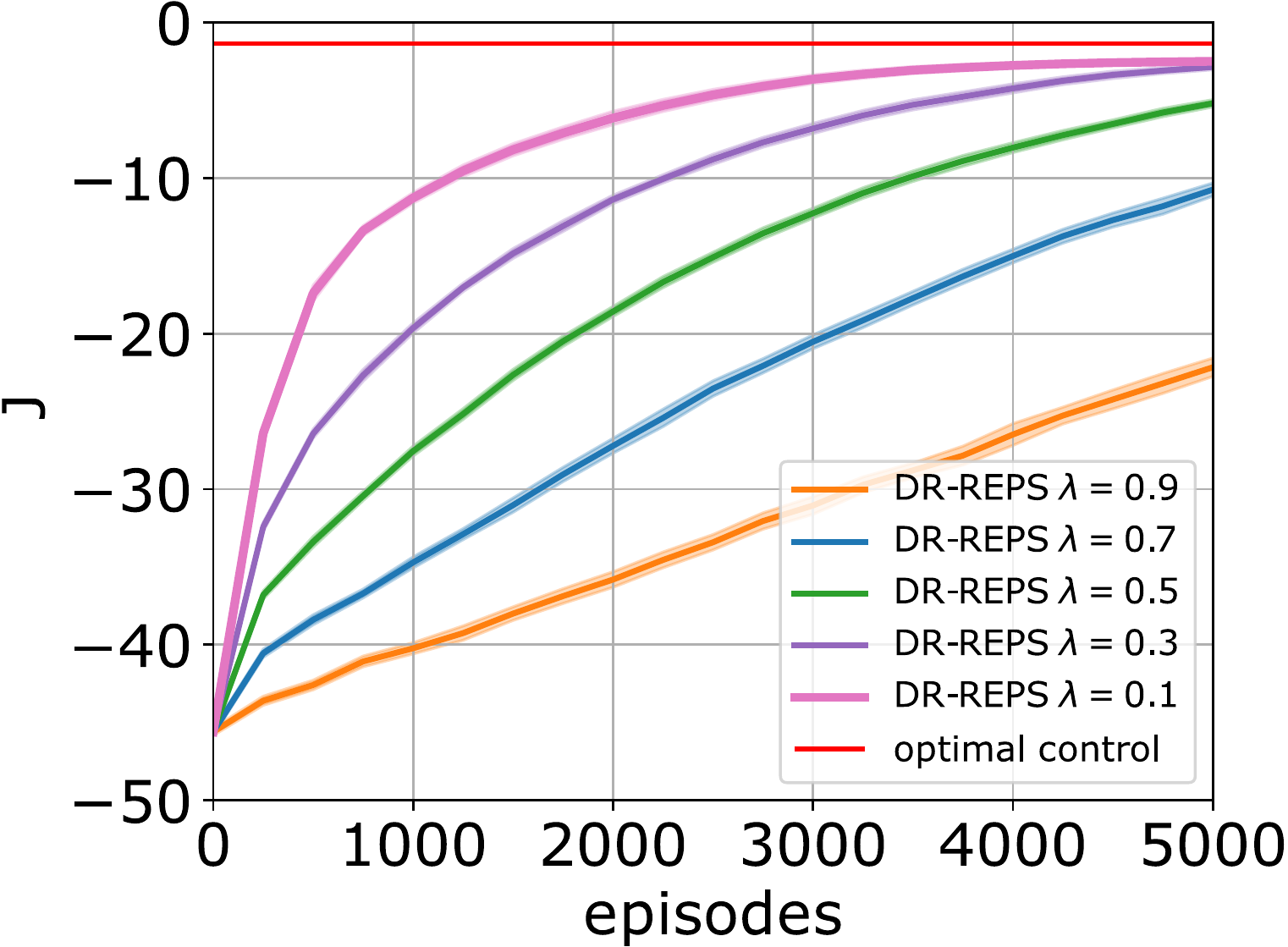}
         \caption{\gls{lqr} - FullCov \\30 selected parameters}
         \label{fig:lqr_ablation_creps_full_k_30}
     \end{subfigure}
     \hfill
     \begin{subfigure}[b]{0.48\textwidth}
         \centering
         \includegraphics[width=\textwidth]{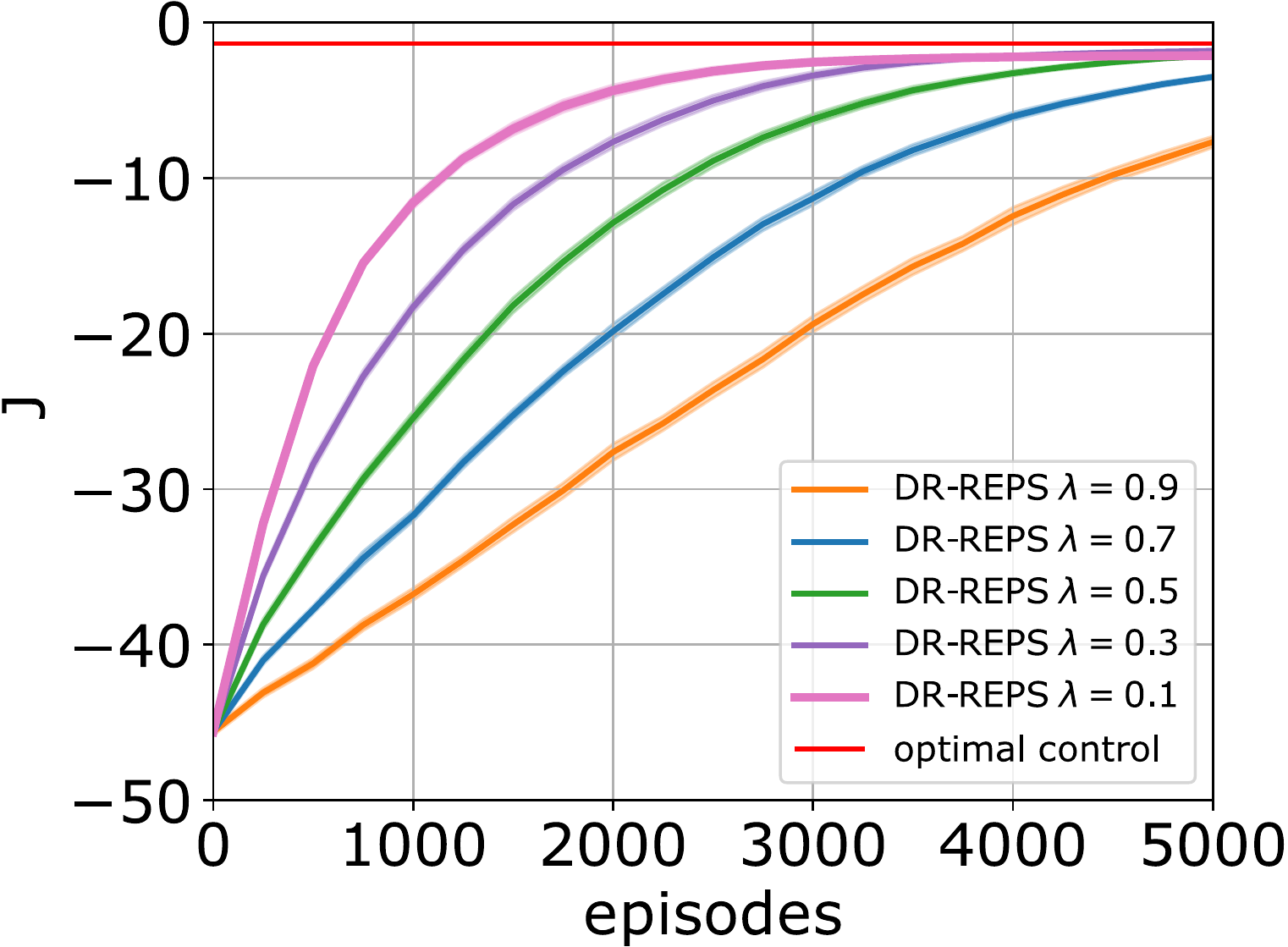}
         \caption{\gls{lqr} - FullCov \\50 selected parameters}
         \label{fig:lqr_ablation_creps_full_k_50}
     \end{subfigure}
        \caption{Ablation study on the \gls{lqr} with the \gls{drreps} algorithm.}
        \label{fig:lqr_ablation_reps}
\end{figure}

\begin{figure}[t]
     \centering
     \begin{subfigure}[b]{0.48\textwidth}
         \centering
         \includegraphics[width=\textwidth]{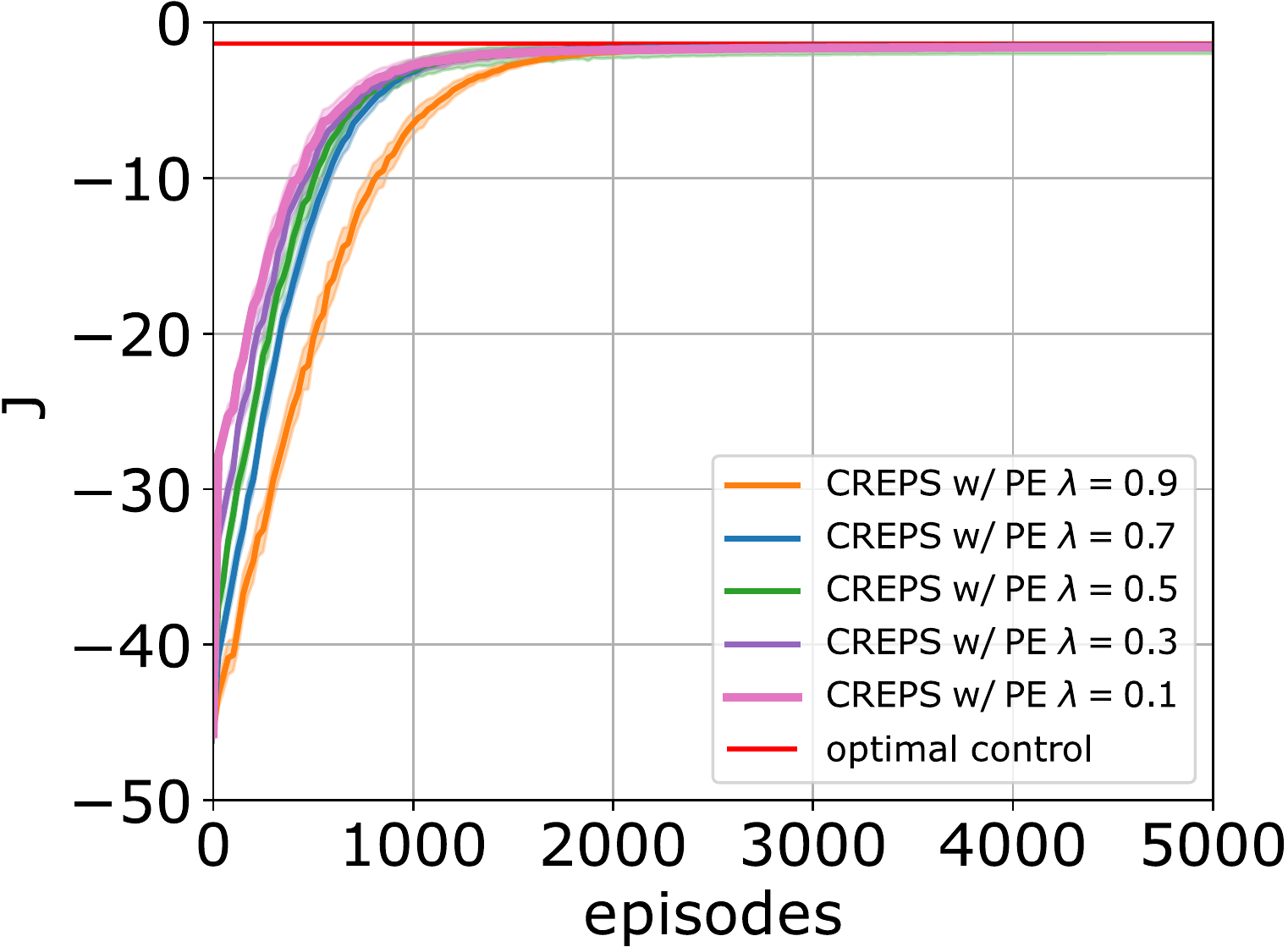}
         \caption{\gls{lqr} - DiagCov \\30 selected parameters}
         \label{fig:lqr_ablation_reps_diag}
     \end{subfigure}
     \hfill
     \begin{subfigure}[b]{0.48\textwidth}
         \centering
         \includegraphics[width=\textwidth]{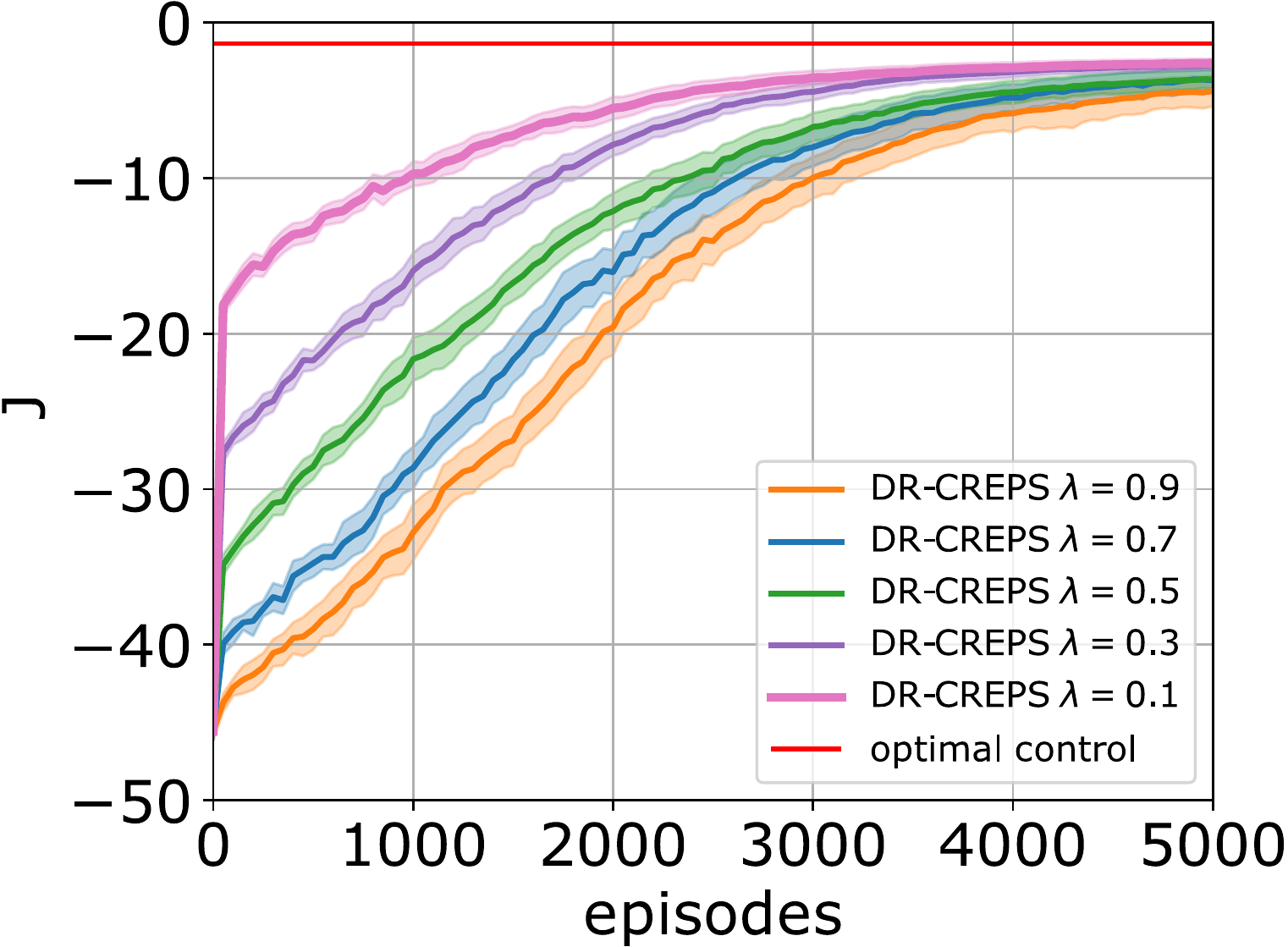}
         \caption{\gls{lqr} - FullCov \\10 selected parameters}
         \label{fig:lqr_ablation_reps_full_k_10}
     \end{subfigure}
     \\
     \vspace{+1em}
     \begin{subfigure}[b]{0.48\textwidth}
         \centering
         \includegraphics[width=\textwidth]{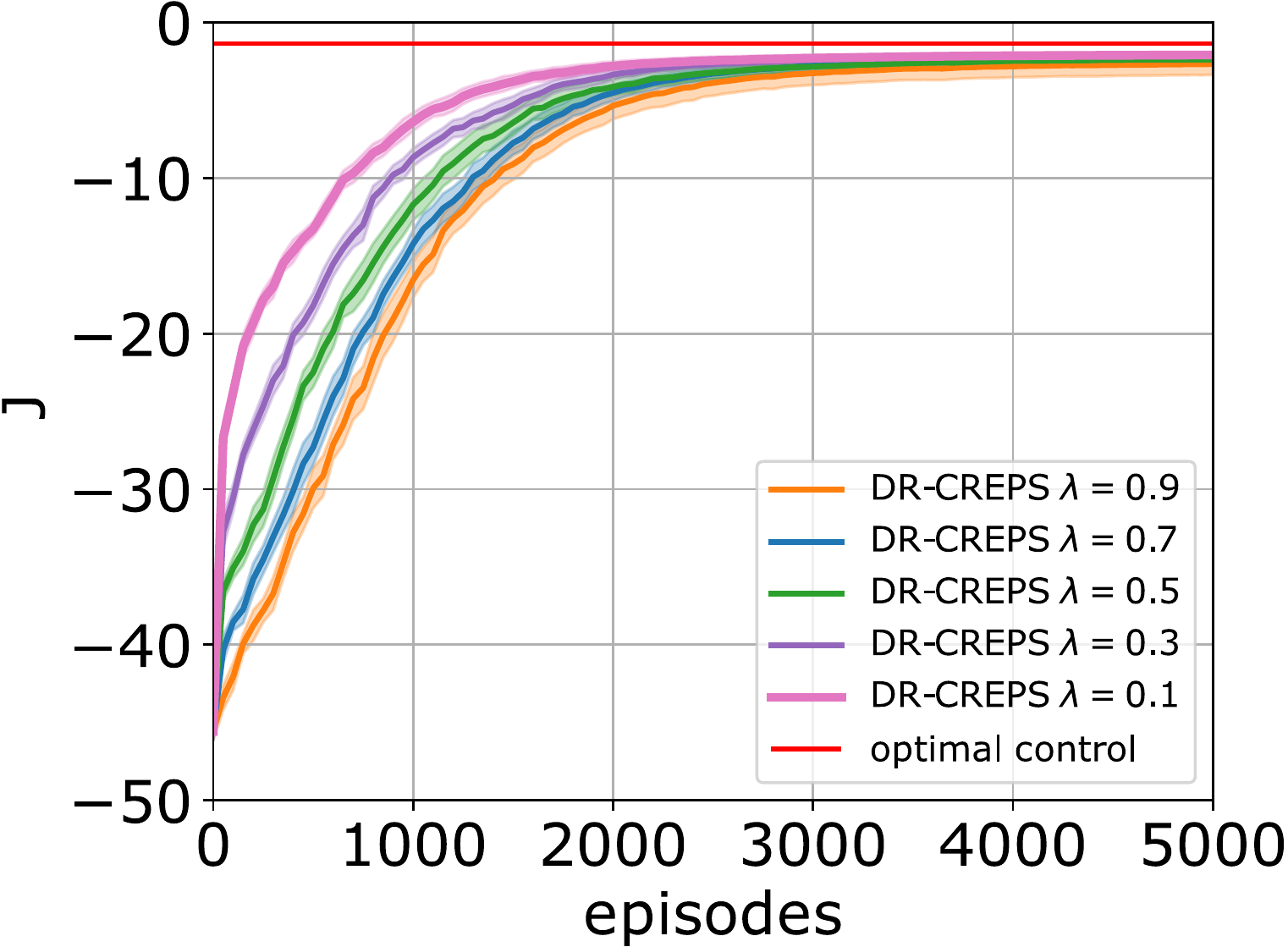}
         \caption{\gls{lqr} - FullCov \\30 selected parameters}
         \label{fig:lqr_ablation_reps_full_k_30}
     \end{subfigure}
     \hfill
     \begin{subfigure}[b]{0.48\textwidth}
         \centering
         \includegraphics[width=\textwidth]{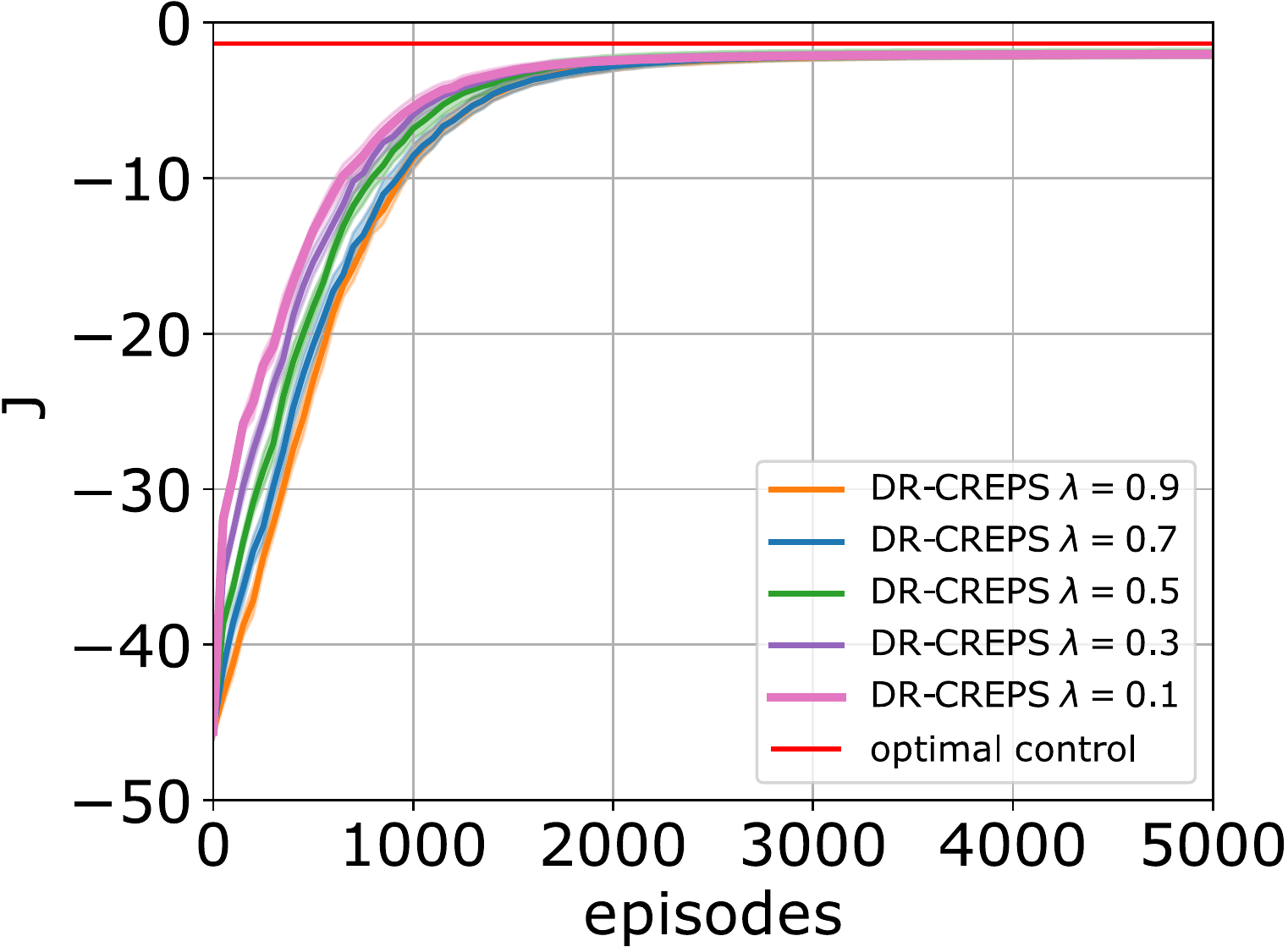}
         \caption{\gls{lqr} - FullCov \\50 selected parameters}
         \label{fig:lqr_ablation_reps_full_k_50}
     \end{subfigure}
        \caption{Ablation study on the \gls{lqr} with the \gls{drcreps} algorithm.}
        \label{fig:lqr_ablation_creps}
\end{figure}

In Fig.~\ref{fig:lqr_ablation_reps} and Fig.~\ref{fig:lqr_ablation_creps} we show the results of our ablation study on the \gls{lqr} environment for the \gls{drreps} and  \gls{drcreps} algorithm, respectively. 
We plot the results for the number of selected parameters separately and vary the value for $\lambda$ inside each plot.

On the diagonal case in Fig.~\ref{fig:lqr_ablation_creps_diag}, hyperparameter $\lambda$ seems to correlate directly with the initial learning speed with significant jumps during the first epochs. This jump comes from focusing the exploration on only the effective parameters and, in the case of $\lambda=0.1$, nearly completely neglecting the ineffective ones by reducing their covariance to a mere 10\% of the original value. In this experiment, $\lambda=0.9$ is too close to the behavior without PE i.e, $\lambda=1.0$, to show a significantly improved learning performance.
These findings get amplified when moving to the full covariance case in Fig.~\ref{fig:lqr_ablation_creps_full_k_10}, Fig.~\ref{fig:lqr_ablation_creps_full_k_30}, and Fig.~\ref{fig:lqr_ablation_creps_full_k_50} when we extend the algorithms with \gls{gdr}. The less effective parameters we select, the larger the initial gain in learning. This gain is because we apply the full step size $\epsilon$ to a subset of the parameters. Especially when selecting only 10 parameters, we observe large jumps at the beginning, which shows that updating only a subset of the parameter, i.e., the effective ones, and focusing the exploration leads to a great boost for the learning process. Selecting fewer parameters, however, comes with the downside of a slightly worse optimum reached. A reason for this might be that too large update steps on too few parameters prevent the algorithm from exploring some effective parameters that would lead to convergence to the optimum.

To prevent premature convergence \gls{creps} introduces the entropy constraint. Since \gls{reps} does not include such, the algorithm experiences premature convergence more easily and we have to be cautious when selecting a value for $\lambda$ that is too close to 0. This is clearly shown in Fig.~\ref{fig:lqr_ablation_reps_diag} which presents the diagonal covariance case on the \gls{lqr}. The closer $\lambda$ is to 0, the larger the initial performance gain but also the loss of final return. Updating only the effective parameters seems again to magnify this tradeoff as visualized in Fig.~\ref{fig:lqr_ablation_reps_full_k_10}, Fig.~\ref{fig:lqr_ablation_reps_full_k_30}, and Fig.~\ref{fig:lqr_ablation_reps_full_k_50}, respectively. In the given number of episodes, $\lambda=0.1$ reaches a better optimum. However, the last episodes of Fig.~\ref{fig:lqr_ablation_reps_full_k_50} hint that higher values might reach a better optimum but require significantly more episodes to do so. This effect is also only really impactful when 50 parameters, i.e., 50\% of the total number of parameters, are selected, which is not feasible for larger dimensional problems as the number of episodes required per update becomes unfeasible.

Finally, our hyperparameter sweeps in Tab.~\ref{tab:hyper_air_hockey_0} and Tab.~\ref{tab:hyper_ball_stopping_0} show that \gls{promp} are also quite sensitive to $\lambda$ and the best performance/learning tradeoff is achieved at $\lambda=0.5$. %\puze{Do we have plot for it?} 
Therefore, we can not always select $\lambda$ as low as possible, but we should search for the best hyperparameter. However, our findings suggest that for a linear regressor $\lambda=0.1$, and \gls{promp}s $\lambda=0.5$ seem like good and mostly sufficient starting points.

\subsection{Comparison between \texorpdfstring{\acrlong{mi}, \acrlong{pearson}}{Lg}, and Random}
\label{sec:corelation_comparison}
\begin{figure}[t]
     \centering
     \begin{subfigure}[b]{0.48\textwidth}
         \centering
         \includegraphics[width=\textwidth]{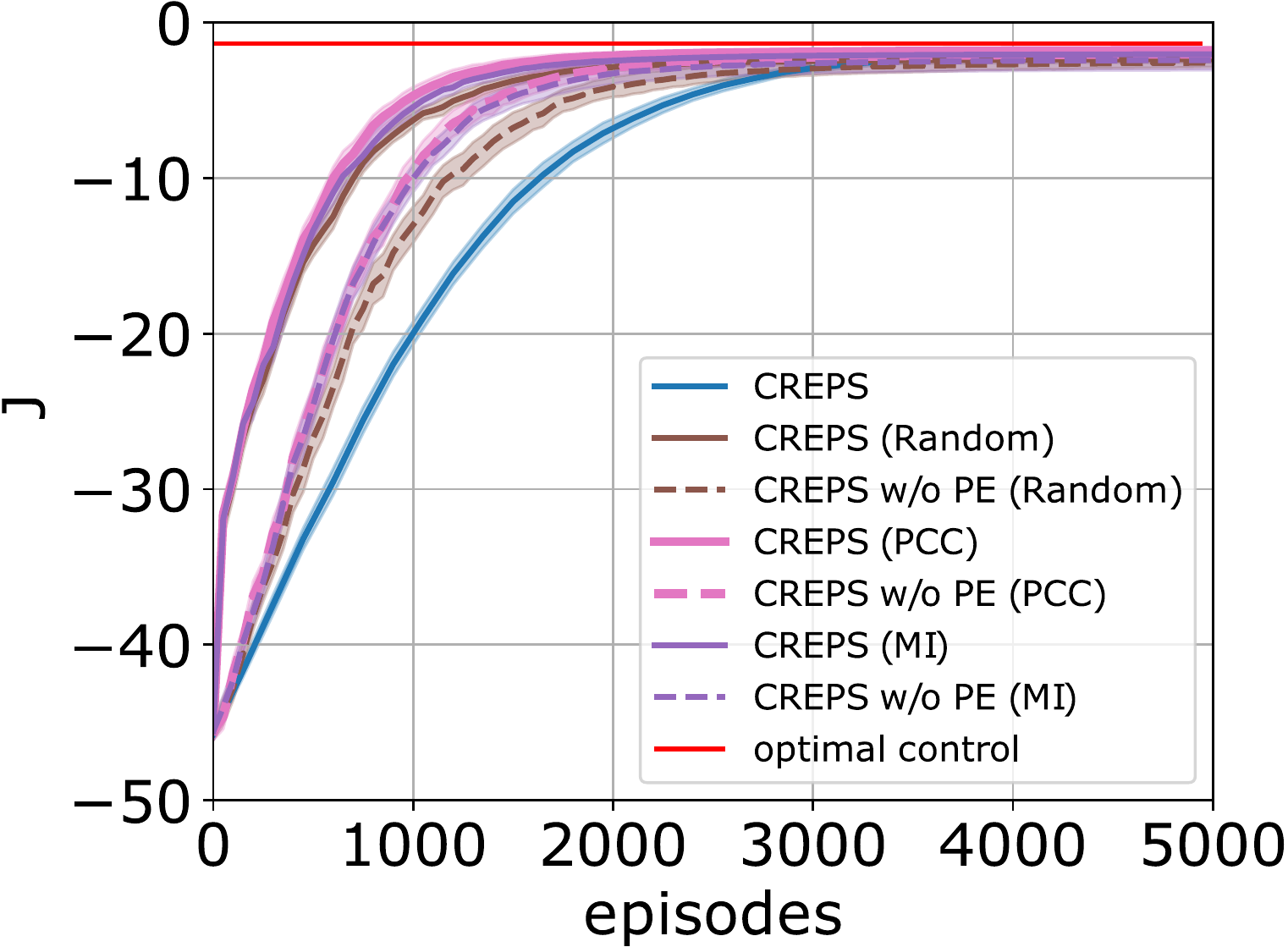}
         \caption{\gls{lqr}}
         \label{fig:lqr_ablation_random}
     \end{subfigure}
     \hfill
     \begin{subfigure}[b]{0.48\textwidth}
         \centering
         \includegraphics[width=\textwidth]{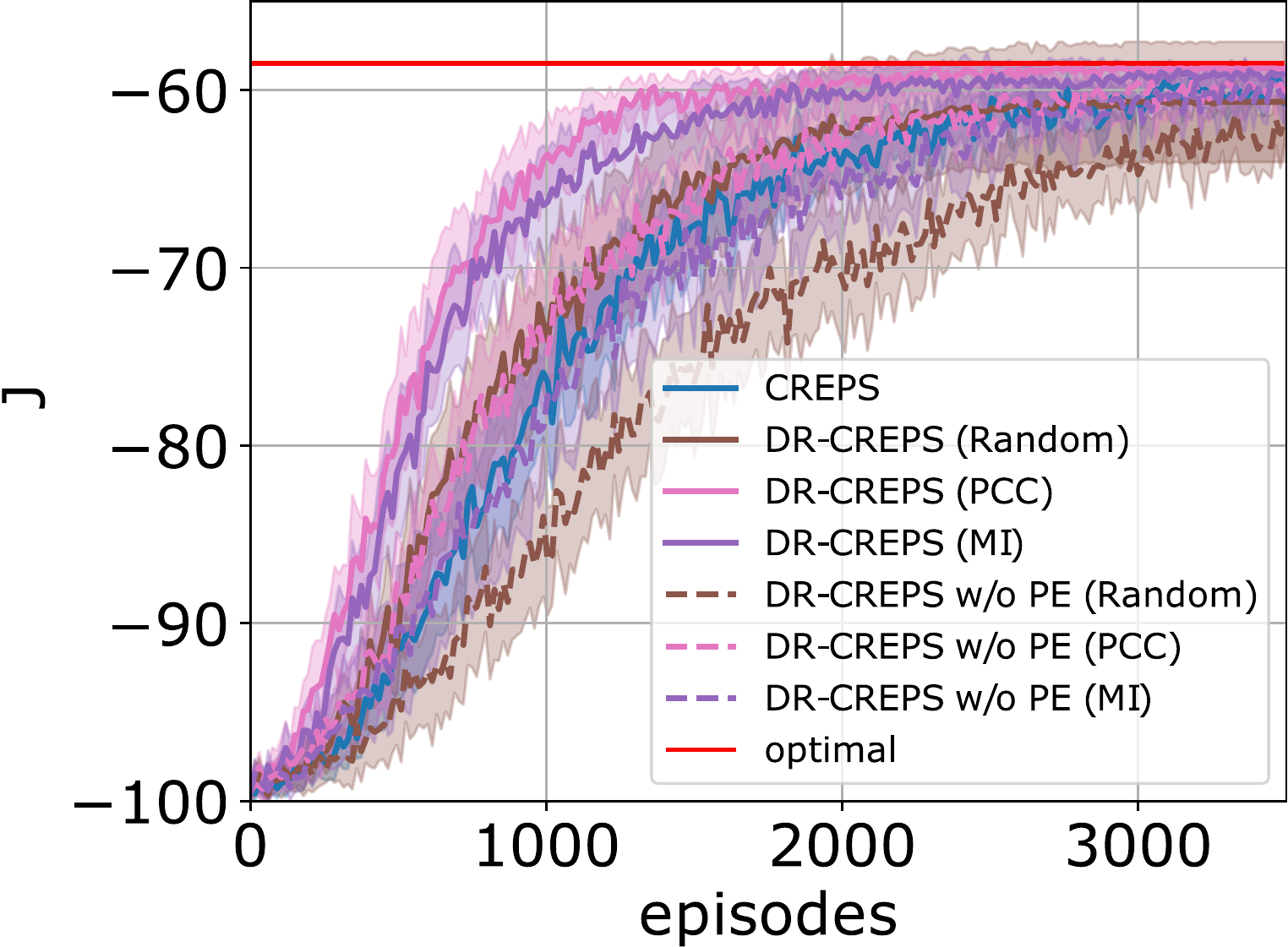}
         \caption{ShipSteering}
         \label{fig:ship_steering_ablation_random}
     \end{subfigure}
     \\
     \vspace{+1em}
     \begin{subfigure}[b]{0.48\textwidth}
         \centering
         \includegraphics[width=\textwidth]{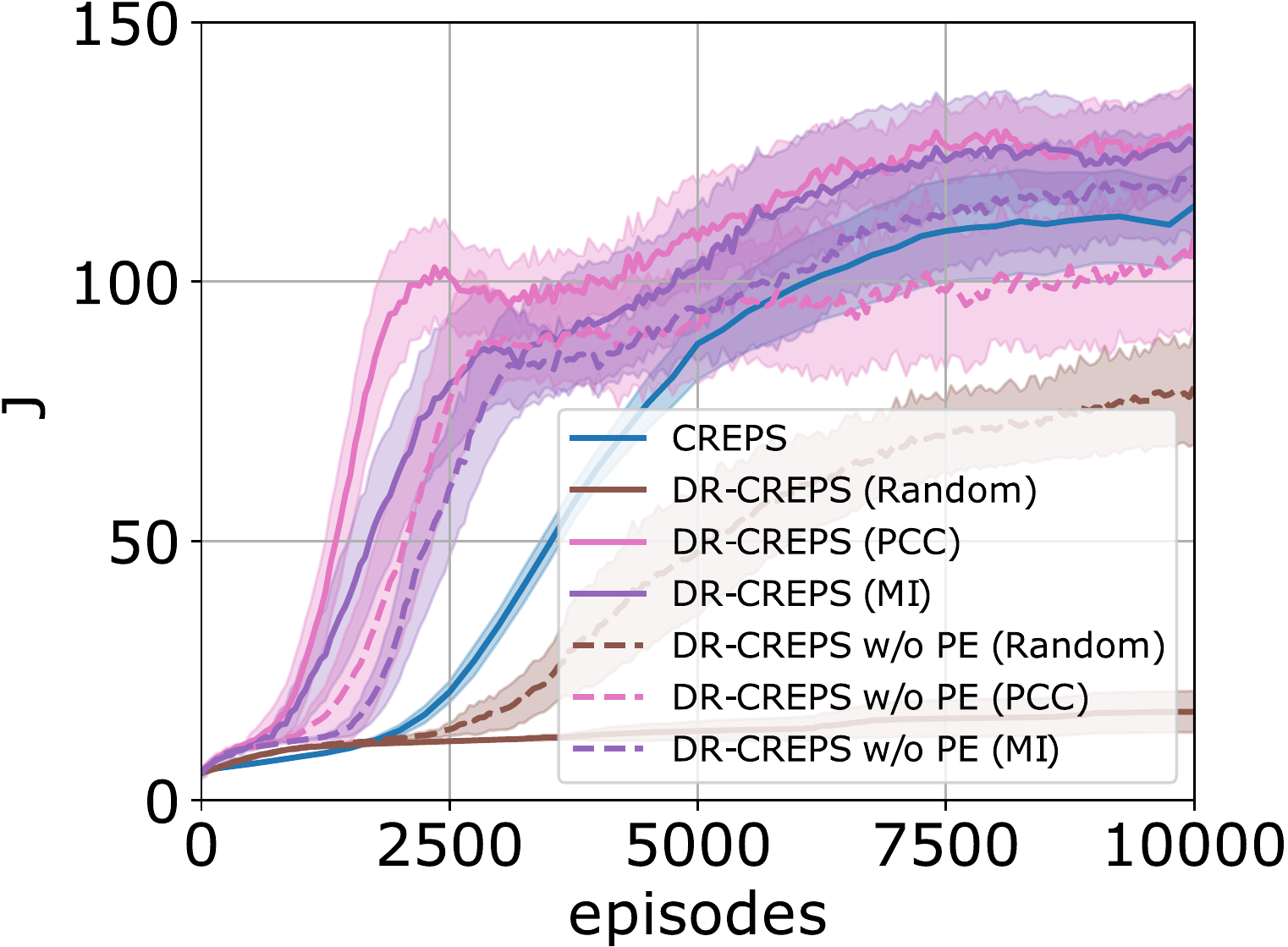}
         \caption{AirHockey}
         \label{fig:air_hockey_ablation_random}
     \end{subfigure}
     \hfill
     \begin{subfigure}[b]{0.48\textwidth}
         \centering
         \includegraphics[width=\textwidth]{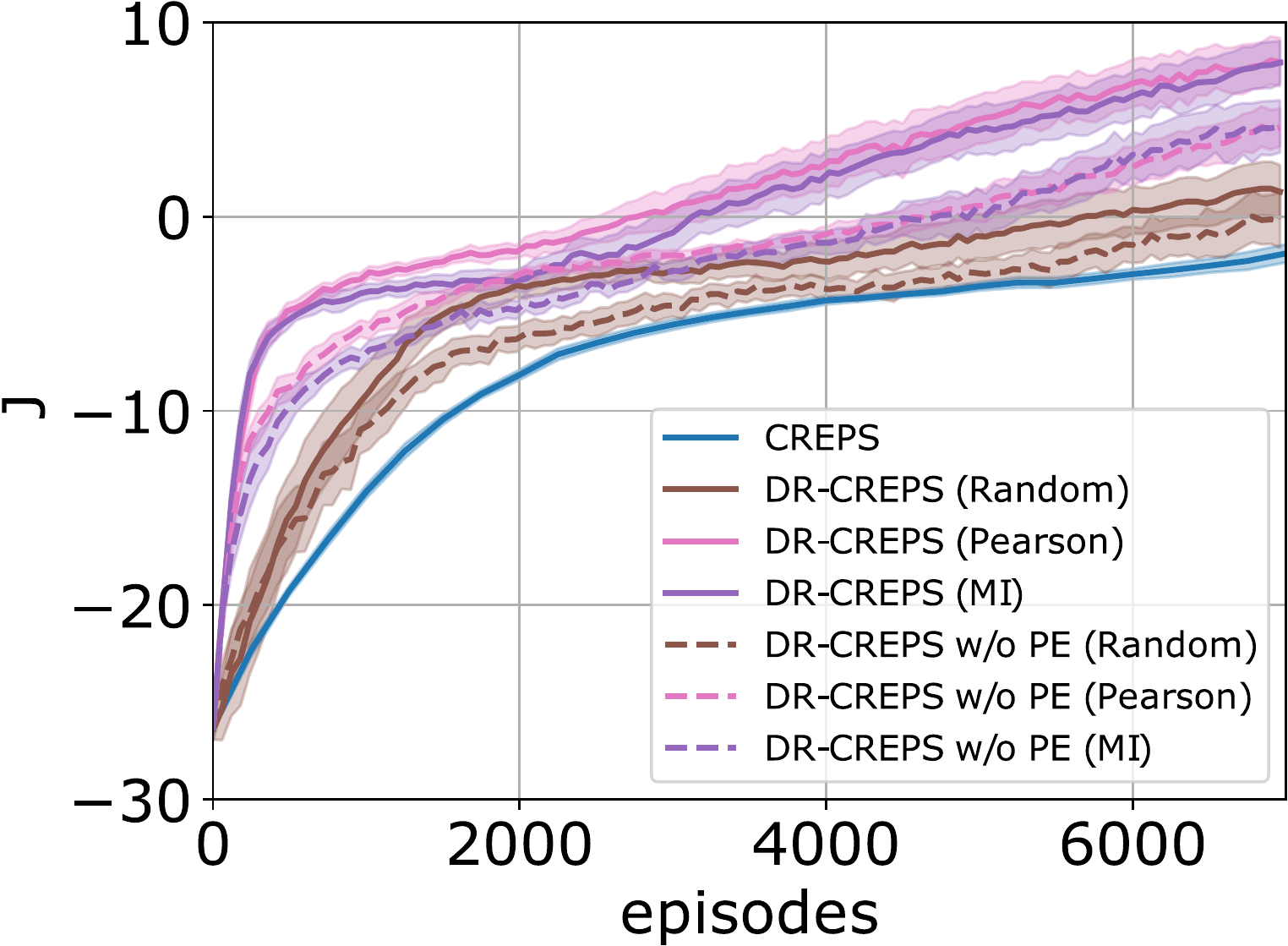}
         \caption{BallStopping}
         \label{fig:ball_stopping_ablation_random}
     \end{subfigure}
        \caption{Comparison of \gls{mi}, \gls{pearson}, and a random selector (Random) to select the effective parameters.}
        \label{fig:all_ablation_random}
\end{figure}

To verify that using \gls{mi} and \gls{pearson} benefits the parameter selection process, we compare it to a random selector that represents the most primitive baseline. We conduct this part of the ablation study in all environments.

The results on the \gls{lqr} shown in Fig.~\ref{fig:lqr_ablation_random} suggest that our method already accounts for some of the performance gains and can outperform the vanilla \gls{creps} in terms of learning speed even when using a random selector. We attribute this to the fact that reducing the covariance of some subset of parameters limits the current exploration space and makes it easier to learn. If some effective parameters are selected using a correlation metric or even randomly, these benefits increase. The former is to be preferred as the correlation metric picks parameters associated with higher rewards. The experiments on the simulated environments support these assumptions.

In \textit{ShipSteering}, the main contribution to the learning speed comes from \gls{pe}. \gls{gdr} does not have a significant impact besides reducing the number of episodes per fit. Updating arbitrary parameters makes the updates noisy and leads to worse behavior than \gls{creps}. The reason for this behavior is the observation space which is transformed via rectangular tilings. Since those are binary, updating a subspace does not provide additional gains as only the parameters corresponding to the activated tiles are updated. \gls{pe}, however, helps to focus the exploration on only the effective parameters which correspond to the tiles required to find a good or even optimal path leading to faster learning. The random selection is not sufficient and provides no guarantee of convergence indicated by the large confidence interval.

The \textit{AirHockey} task shows a high instability of both the random selector as well as \gls{gdr} without exploration steering noticeable on the high confidence interval. We further document failure rates due to a violation of the positive definiteness of the covariance matrix of 16\% \gls{drcreps} w/o \gls{pe} (MI), 48\% \gls{drcreps} w/o \gls{pe} (PCC), 44\% \gls{drcreps} w/o \gls{pe} (Random), and 24\% \gls{drcreps} w/o \gls{pe} (Random). This result is similar to the one in \textit{ShipSteering} where \gls{pe} allows us to run larger update steps using fewer episodes. To reduce the failure rate it's possible to either decrease $\epsilon$ or increase the number of episodes per update. Since both solutions would perturb the comparison, we stick to reporting the failure rates instead. Nonetheless, the random selector shows this faulty behavior even when \gls{pe} is applied.

The results on the \textit{BallStopping} task in Fig.~\ref{fig:ball_stopping_ablation_random} support the findings on the \gls{lqr} showing that the correlation measure methods outperform the random selector, even if the latter improves the learning process. 

In conclusion, the above-mentioned experiments show that using the correlation measures as inductive bias significantly improves the performance over a random selector.

\section{MUTUAL INFORMATION ESTIMATION}
\label{sec:mutual_info_estimation}

We investigate three ways to estimate the mutual information. Therefore, we construct a toy example with two random variables $\bm{X},\bm{Y}$ and noise $\bm{E}$ all Gaussian distributed which allow for an analytical estimation of the \gls{mi}. According to \cite{bishop2006pattern} exploiting the properties of multivariate Gaussian distributions, we get the following marginal, joint, and conditional distributions which are also Gaussian distributed:

\begin{align*}
    p(\bm{E}) & = N(\bm{e}|\bm{\mu_e, \Sigma_e}) \\
    p(\bm{X}) &= N(\bm{x}|\bm{\mu_{xx}}, \bm{\Sigma_{xx}}) \\
    p(\bm{Y}|\bm{X}) &= N(\bm{y}|\bm{A}\bm{x}+\bm{\mu_e}, \bm{A}\bm{\Sigma_{xx}}\bm{A}^{T} + \bm{\Sigma_e}) \\
    p(\bm{Y}) &= N(\bm{y}|\bm{A}\bm{\mu_{xx}}, \bm{\Sigma_{yx}}+\bm{A}\bm{\Sigma_{xx}}\bm{A^T}) \\
    p(\bm{X}|\bm{Y}) &= N(\bm{x}|\bm{\Sigma_{xy}}\{\bm{A^T}\bm{\Sigma_{yx}^{-1}}(\bm{y}-\bm{\mu_e})+\bm{\Sigma_{xx}^{-1}}\bm{\mu_{xx}}\},\bm{\Sigma_{xy}}) \\
    & where \quad \bm{\Sigma_{xy}} = (\bm{\Sigma_{xx}^{-1}}+\bm{A^T}\bm{\Sigma_{yx}^{-1}}\bm{A})^{-1}
\end{align*}

If $\bm{X}$ has dimensionality $M$ and $\bm{Y}$ has dimensionality $N$, and $\bm{E}$ has dimensionality $N$, $\bm{A}$ denotes a transformation matrix of full rank with dimensionality $N\times M$.
In this case $\bm{A},\bm{\mu_e},\bm{\mu_{xx}}$ are the parameters governing the means.
We compute the \gls{mi} analytically as well as through three sample-based estimators. 

First, we take the trivial approach of estimating the \gls{mi} through its probabilistic formulation referenced in Eq.~\ref{eq:mi} which we refer to as $\bm{MI_{histogram}}$. We use histograms to estimate the densities of the underlying probability distributions and subsequently compute their entropy $H(\cdot)$. The relation of the entropy to the \gls{mi} is as follows:

\begin{align*}
    MI[\bm{X};\bm{Y}] & = H(\bm{X}) - H(\bm{X}|\bm{Y}) \\
    & = H(\bm{Y}) - H(\bm{Y}|\bm{X}) \\
    & = H(\bm{X}) + H(\bm{X}) + H(\bm{X},\bm{Y}) \\
    & = H(\bm{X},\bm{Y}) - H(\bm{X}|\bm{Y}) - H(\bm{Y}|\bm{X})
\end{align*}

Second, we turn to the estimator proposed by \textit{Kraskov, Stögbauer, Grassberger} (KSG) \citep{Kraskov2004} referred to as $\bm{MI_{KSG}}$.
Lastly, we evaluate scikit-learn's mutual information regressor~\citep{scikit-learn} referred to as $\bm{MI_{regression}}$. The implementation utilizes a combination of \citep{Ross2014, Kraskov2004} which are both based on \citep{Kozachenko1987}.

We choose hyperparameter $bins$ per histogram ($MI_{histogram}$) and $k$ nearest neighbors ($MI_{regression}$, $MI_{ksg}$) to reflect our goal of using fewer samples per update and estimation step. Both parameters must be less than the number of samples used for the estimation. We found that $MI_{histogram}$ is more sensitive to $bins$ while $MI_{regression}$ and $MI_{ksg}$ produce similar results with varying $k$. We choose values of $bins=4$ and $k=4$ as a tradeoff of bias and performance~\citep{scikit-learn}.

For visualization purposes, we evaluate the estimators in the simplified case of $M=1, N=1$ and compute results for 4 rollouts. For each rollout, we randomize the Gaussian noise $p(\bm{E})$, sampling process $\bm{X}\sim p(\bm{x}), \bm{E}\sim p(\bm{E})$, and transformation matrix $\bm{A}$ while using the same samples from $p(\bm{X})$. 

\begin{figure}[t]
     \centering
     \begin{subfigure}[b]{0.48\textwidth}
         \centering
         \includegraphics[width=\textwidth]{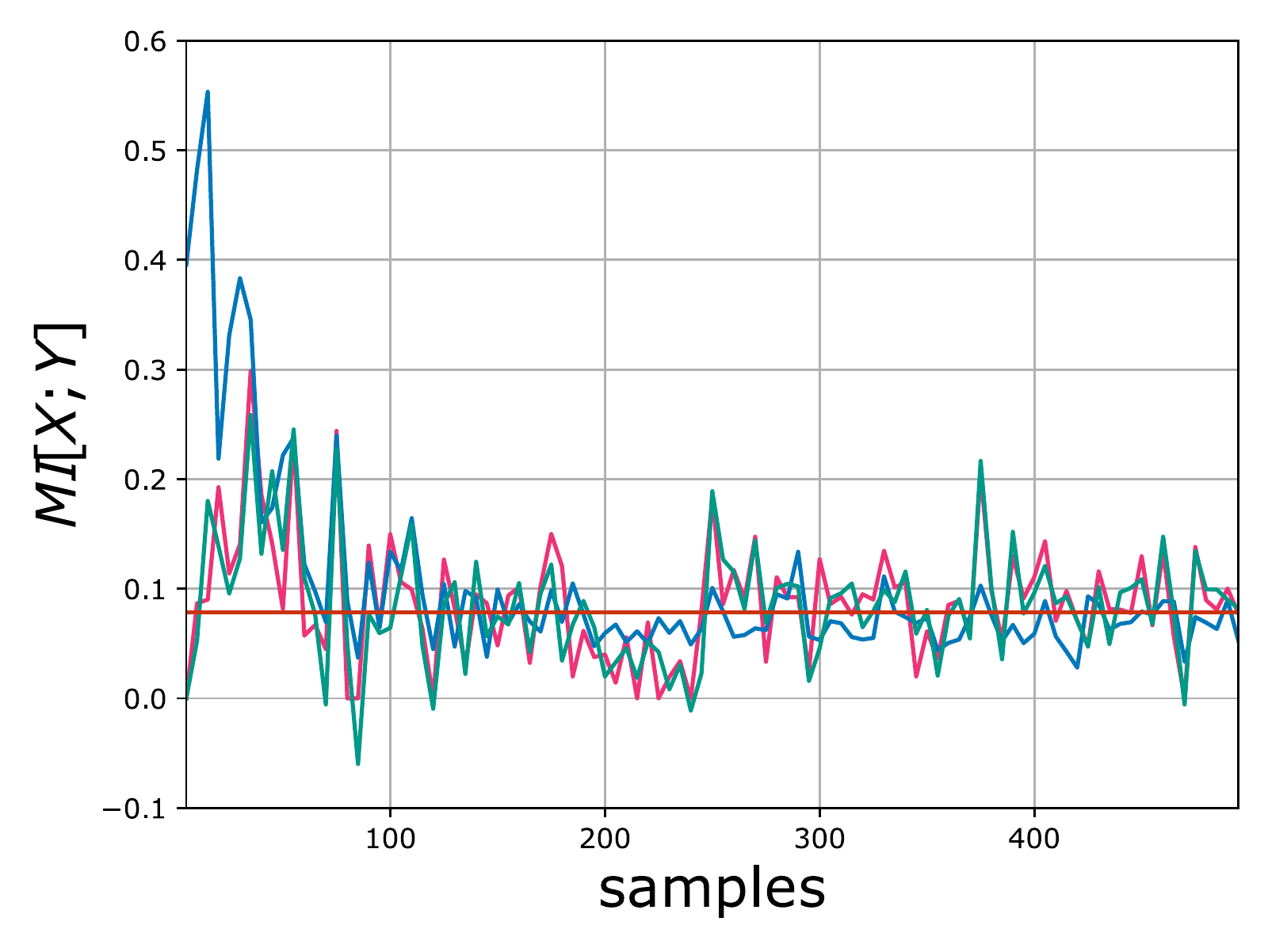}
        \caption{Random seed: 0}
        \label{fig:mi_estimators_0}
     \end{subfigure}
     \hfill
     \begin{subfigure}[b]{0.48\textwidth}
         \centering
         \includegraphics[width=\textwidth]{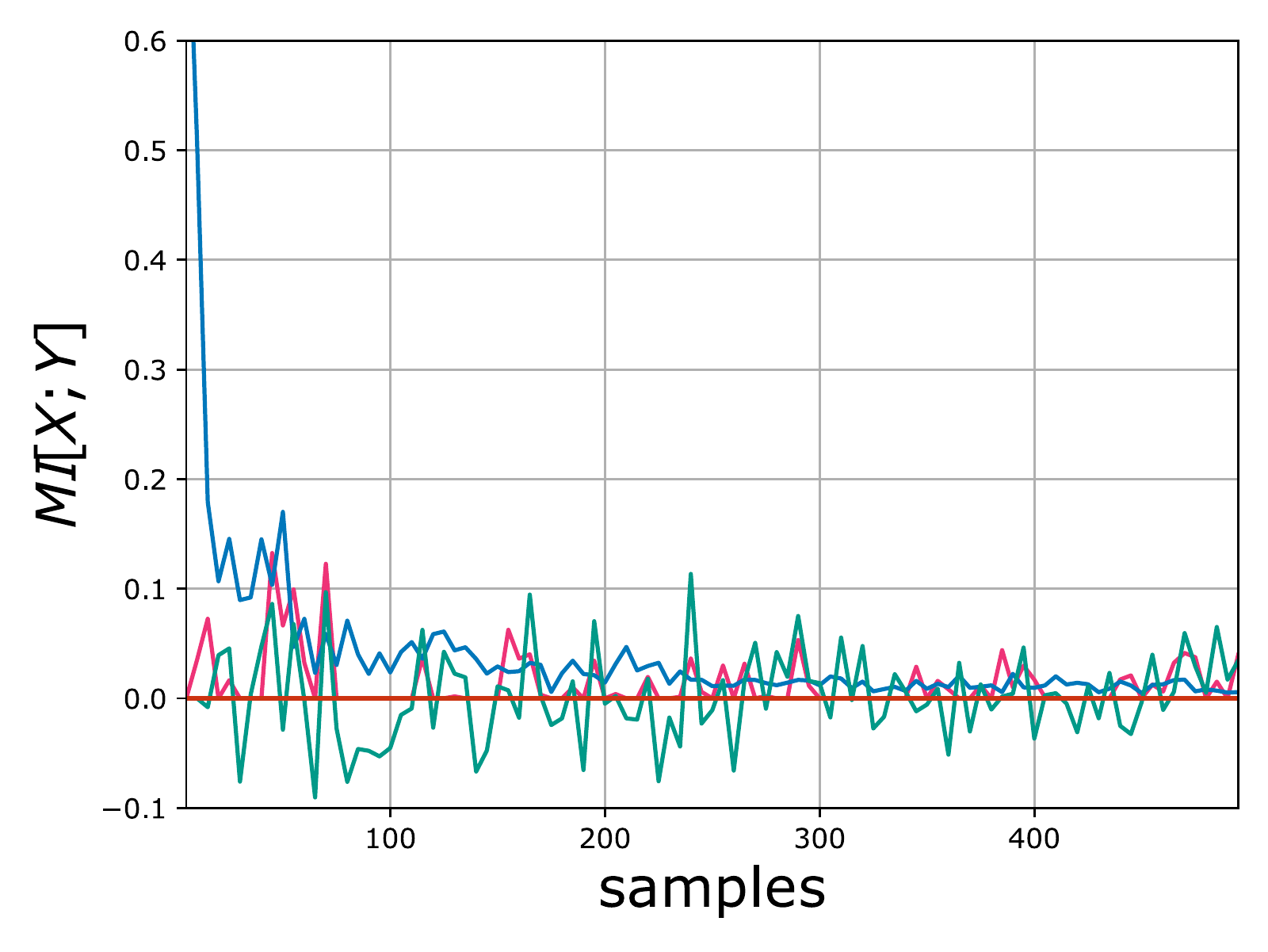}
     \caption{Random seed: 1}
     \label{fig:mi_estimators_1}
     \end{subfigure}
     \\
     \vspace{+1em}
    \begin{subfigure}[b]{0.48\textwidth}
        \centering
        \includegraphics[width=\textwidth]{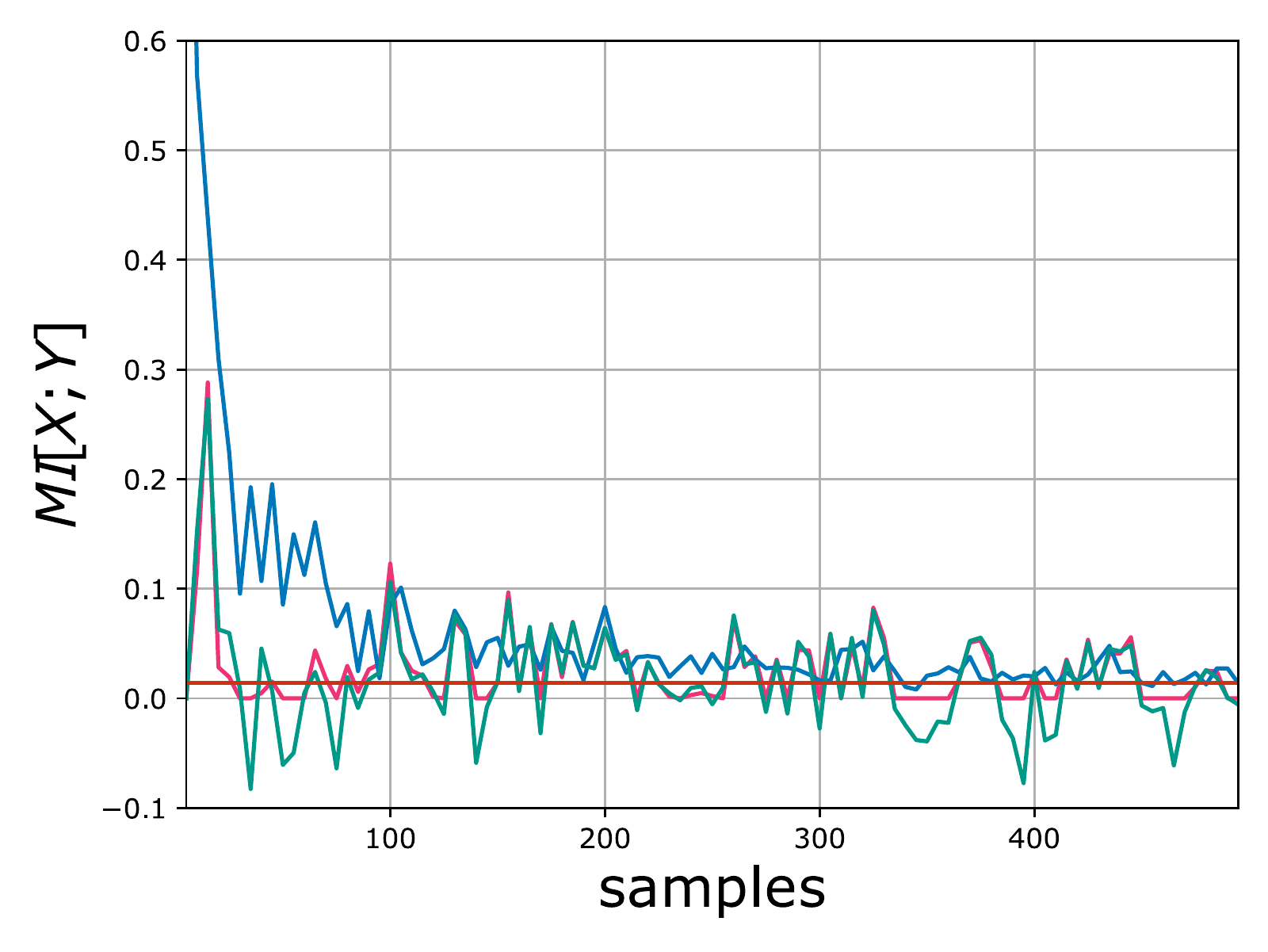}
     \caption{Random seed: 2}
     \label{fig:mi_estimators_2}
     \end{subfigure}
     \begin{subfigure}[b]{0.48\textwidth}
        \centering
        \includegraphics[width=\textwidth]{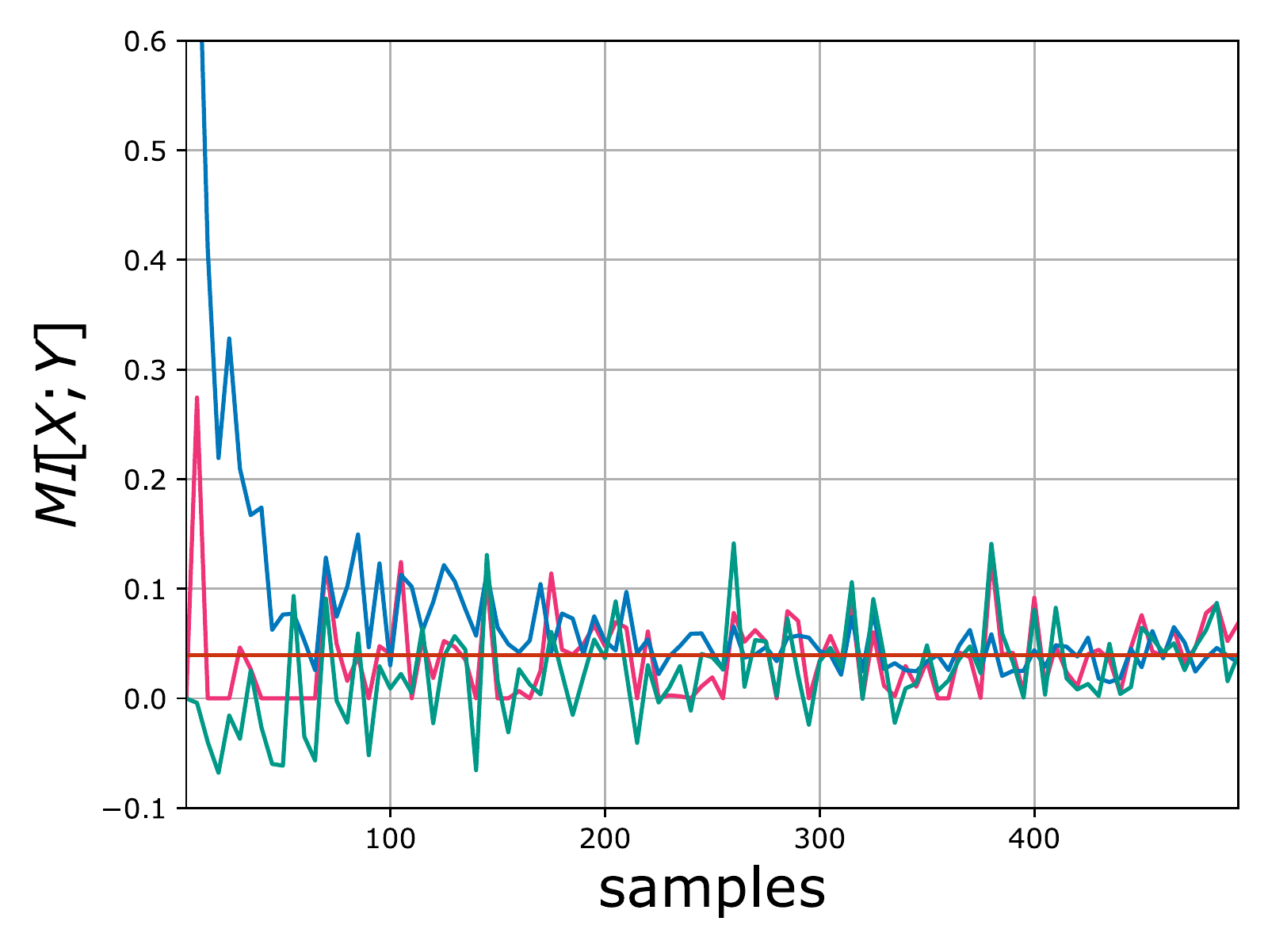}
     \caption{Random seed: 3}
     \label{fig:mi_estimators_3}
     \end{subfigure}
    \begin{subfigure}[b]{0.7\textwidth}
         \vspace{1em}
         \centering
         \includegraphics[width=\textwidth]{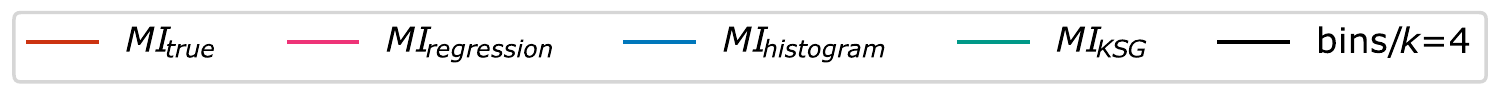}
     \end{subfigure}
     \caption{Comparison of different \gls{mi} estimators on our toy problem over multiple random seeds.}
     \label{fig:mi_estimators}
\end{figure}

Fig.~\ref{fig:mi_estimators} shows the results of this experiment. Since each dimension experiences a different transformation and noise the analytical solutions and estimates for the \gls{mi} vary between them. The $MI_{histogram}$ based approach requires far more samples to achieve a stable estimate of the \gls{mi} compared to the other methods. As expected, $MI_{regression}$ and $MI_{ksg}$ show similar behavior because they use the same underlying estimator. However, $MI_{regression}$ is limited to estimates $\geq 0$ and standardizes the samples. These changes lead to a more stable estimation especially in the low samples regime of $0-200$ which is similar to the range we use for our \gls{drcreps} experiments. These findings support our decision of using the $MI_{regression}$ estimator with hyperparameter $k=4$ for all experiments in the main paper.

\end{document}